\documentclass[final]{nextgame2026}

\usepackage{booktabs}
\usepackage{subcaption}
\usepackage{tikz}
\usetikzlibrary{arrows.meta,positioning,shapes.geometric,fit,calc,backgrounds}
\usepackage[dvipsnames]{xcolor}
\usepackage{multirow}
\usepackage{array}
\usepackage{bm}

\usepackage{titlesec}
\titlespacing*{\section}{0pt}{8pt plus 2pt minus 2pt}{4pt plus 1pt minus 1pt}
\titlespacing*{\subsection}{0pt}{6pt plus 2pt minus 2pt}{3pt plus 1pt minus 1pt}
\titlespacing*{\paragraph}{0pt}{4pt plus 1pt minus 1pt}{4pt}

\title[Poker Arena]{Poker Arena: Multi-Axis Profiling of Strategic Reasoning and Memory in LLMs}

\optauthor{%
    \Name{Pratham Singla} \Email{pratham\_s@me.iitr.ac.in}\\
    \addr Indian Institute of Technology Roorkee
    \AND
    \Name{Shivank Garg} \Email{shivank\_g@mfs.iitr.ac.in}\\
    \addr Indian Institute of Technology Roorkee
    \AND
    \Name{Vihan Singh} \Email{vihan@raeth.ai}\\
    \addr Raeth AI
} 
\usepackage[most,breakable]{tcolorbox}

\newtcblisting{promptbox}[1]{
  enhanced, breakable, size=small,
  colback=black!3!white, colframe=RoyalBlue!70!black,
  fonttitle=\bfseries\small\sffamily, title={#1},
  boxrule=0.5pt, arc=2pt,
  left=4pt, right=4pt, top=4pt, bottom=4pt,
  listing only,
  listing options={
    basicstyle=\footnotesize\ttfamily,
    breaklines=true, columns=fullflexible,
    keepspaces=true, showstringspaces=false,
    aboveskip=0pt, belowskip=0pt
  }
}

\begin{document}

\maketitle

\begin{abstract}
Strategic reasoning under uncertainty underpins consequential decisions in negotiation, finance, and policy, but prevailing game-play benchmarks collapse heterogeneous reasoning dimensions into a single scalar, leaving the capability structure of frontier LLMs unexamined. We introduce \textit{Poker Arena}, a no-limit Texas Hold'em tournament platform that couples a three-layer memory architecture (within-hand, session, and cross-session) with a nine-axis cognitive profile decomposing strategic reasoning into interpretable dimensions such as bet-sizing calibration and positional awareness. We evaluate seven frontier models across 50 sessions of 1{,}000 hands and a controlled memory ablation; tournament chips and aggregate axis score order the field differently: Claude Opus 4.6 wins $+\$15{,}730$ chips with 14 first-place finishes, yet ranks only fifth of seven on mean axis score, while persistent memory helps some models and hurts others. These findings show that multi-axis evaluation surfaces capability structure that scalar leaderboards systematically misrank, with cross-dimensional consistency outweighing peak performance on any single axis.

\end{abstract}

\section{Introduction}\label{sec:intro}

Frontier large language models (LLMs) are increasingly expected to plan and act under partial observability against adaptive opponents, conditions that put strategic reasoning under uncertainty at the centre of contemporary LLM evaluation. Poker is a particularly dense instantiation: within a single decision an agent must size a bet, decide whether to bluff, model heterogeneous opponents, and reason about expected value under stochastic run-outs. These competencies are dissociable, which makes poker a natural testbed for decomposing strategic reasoning rather than reducing it to a single outcome.

Two challenges set the evaluation of frozen-weight LLMs in poker apart from classical poker AI. State-of-the-art poker agents rely on counterfactual regret minimization and large-scale self-play search \citep{bowling2015hulhe,brown2018libratus,brown2019pluribus}, which are training-time methods inapplicable at inference; conversely, naive prompt-based agents lack persistence across hands, discarding the historical signal strategic adaptation depends on. Existing memory architectures \citep{packer2023memgpt,shinn2023reflexion,park2023generative} have been studied primarily in single-agent or cooperative settings, leaving competitive multi-agent behaviour largely uncharacterised, while game-play benchmarks \citep{meta2022cicero,duan2024gtbench} report aggregate metrics that conflate distinct strategic competencies and rarely treat persistent memory as a controlled variable. A full survey of related work (LLM game-play benchmarks, classical poker AI, opponent modelling, persistent-memory architectures, and theory-of-mind evaluations) is in Appendix~\ref{app:related}.

\begin{figure}[!t]
  \centering
  \includegraphics[width=0.85\textwidth]{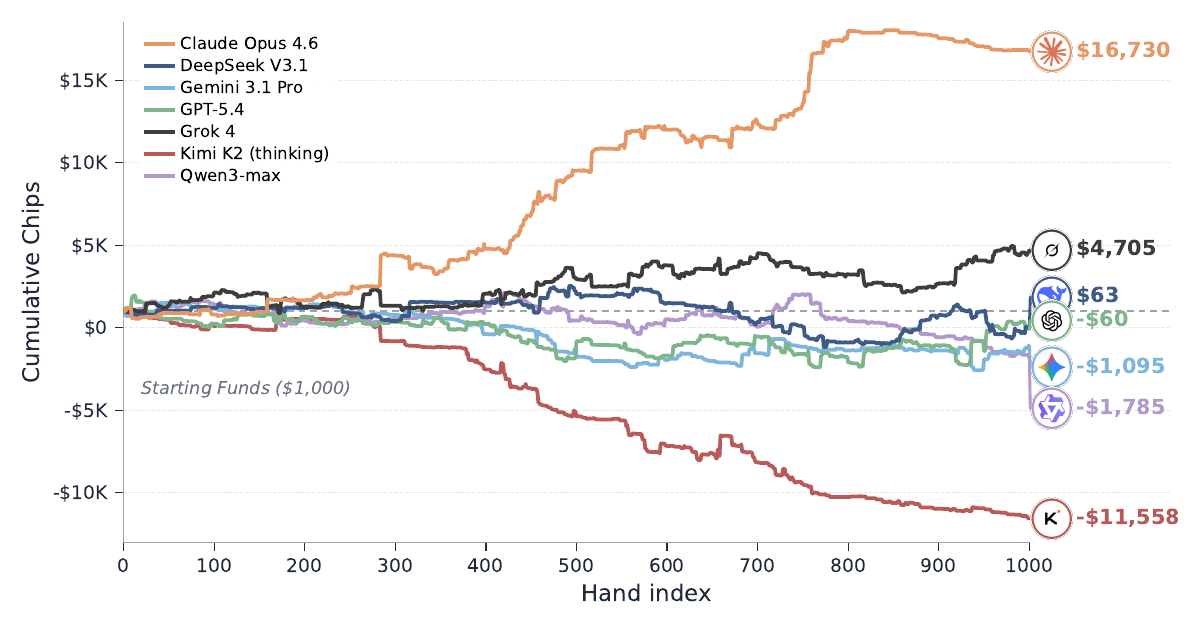}
  \caption{Cumulative chip totals across 1{,}000 hands (50 sessions $\times$ 20 hands, seven seats per session, \$1{,}000 buy-in). Claude finishes at \$16{,}730, roughly four times Grok's \$4{,}705 second-place margin; GPT, DeepSeek, Gemini, and Qwen settle within a \$2{,}000 band of the buy-in line; Kimi trails at $-\$11{,}558$.}
  \label{fig:chip_growth}
\end{figure}

We address these gaps with Poker Arena, an end-to-end no-limit Texas Hold'em tournament platform pairing a controllable three-layer memory system (within-hand context, model-controlled session notebook, and cross-session lifetime store) with a nine-axis profile of strategic competence spanning bet sizing, bluffing, opponent reading, composure, adaptability, prediction, strategic mixing, factual accuracy, and positional awareness. Across 50 seven-player tournaments comprising 1{,}000 hands and a controlled memory ablation (Figure~\ref{fig:chip_growth}), the tournament winner ranks fifth on mean axis score and persistent memory improves some models while degrading others, both showing that consistent cross-axis competence and model-specific memory interfaces matter more than scalar leaderboard rank.
We make the following major contributions:
\begin{itemize}
    \item \textbf{Poker Arena platform.} An end-to-end no-limit Texas Hold'em tournament environment that logs actions, reasoning, and memory updates across seven frontier LLMs, with deterministic side-pot, timeout, and showdown handling.
    \item \textbf{Three-layer memory architecture.} A within-hand, session, and lifetime memory system in which Layer 2 is written by the agent itself and Layer 3 re-seeds subsequent sessions, enabling targeted ablations over the temporal scale of persistence.
    \item \textbf{Nine-axis cognitive profile and empirical study.} A profile that decomposes strategic reasoning into nine interpretable axes spanning deterministic and judge-scored measures, paired with a 1{,}000-hand evaluation across seven frontier models showing that scalar leaderboards misrank competence and that memory effects are model-dependent.
\end{itemize}

\section{Poker Arena Framework}\label{sec:framework}

Poker Arena evaluates LLM agents in no-limit Texas Hold'em tournaments and records actions, reasoning traces, and memory updates. The framework comprises four components: a game engine, a three-layer memory system, a prompt pipeline, and a model gateway over seven frontier models.

\subsection{Game Engine}
Each session comprises twenty hands seated with up to seven agents, with all agents beginning at \$1{,}000 in chips. Blinds escalate across four tiers every five hands (\$5/\$10, \$10/\$20, \$25/\$50, \$50/\$100) to introduce tournament pressure as effective stack depth shortens. 

Hand rankings are evaluated by a deterministic five-from-seven solver, and win probabilities at each decision point are estimated via Monte Carlo run-out sampling. Side pots are tracked through per-player contribution accounting, ensuring correct payoff allocation under unequal stack sizes and preserving fidelity to standard no-limit tournament rules.

\subsection{Three-Layer Memory}

\begin{figure}[!tb]
  \centering
  \includegraphics[width=0.85\linewidth]{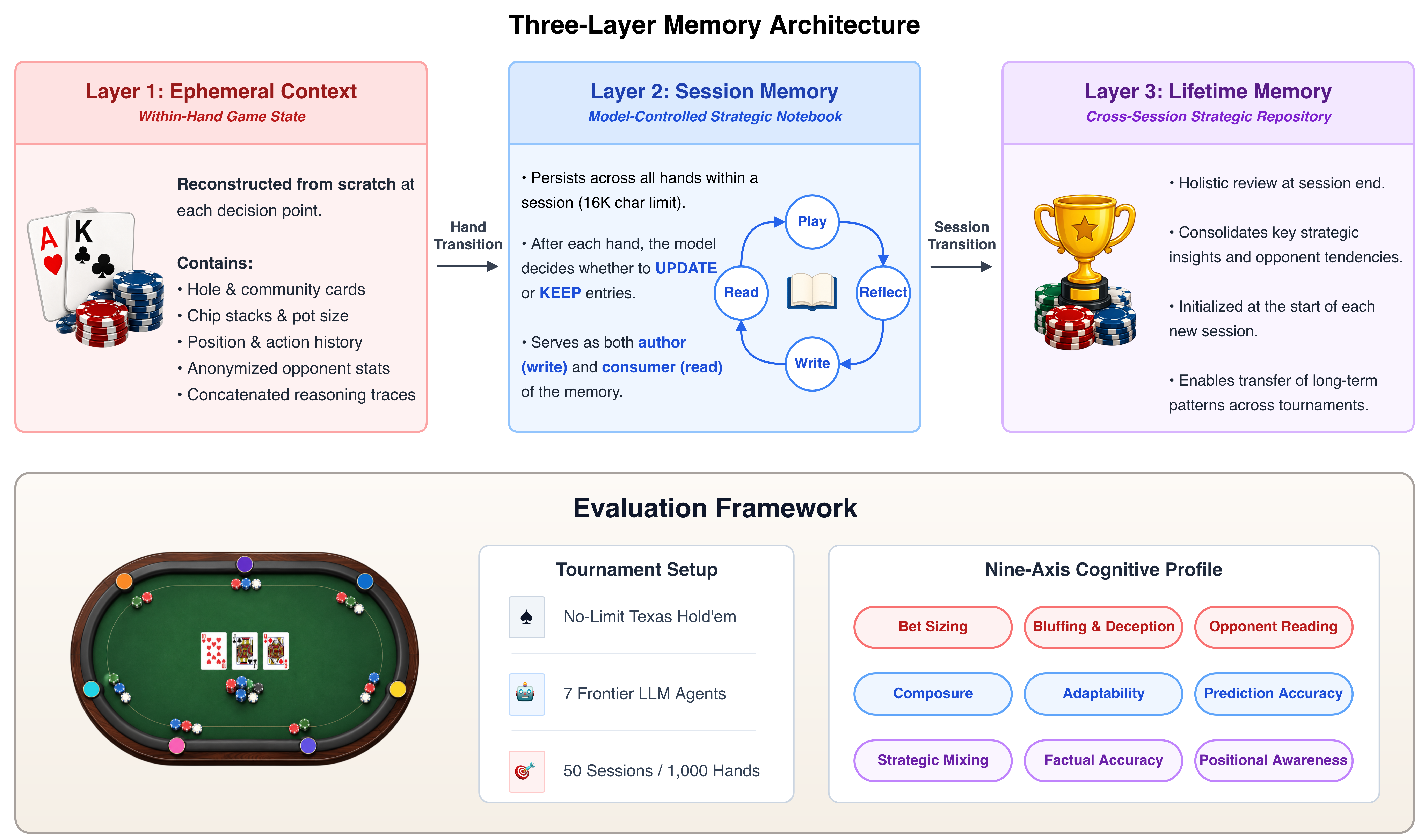}
  \caption{Poker Arena architecture: a three-layer memory hierarchy (Layer~1 within-hand context, Layer~2 session notebook, Layer~3 cross-session lifetime store) feeding seven frontier LLMs over a deterministic Texas Hold'em engine, scored on a nine-axis cognitive profile.}
  \label{fig:memory}
\end{figure}

Strategic reasoning in poker spans multiple temporal scales, including within-hand inference, across-hand adaptation, and cross-session learning. To capture these dynamics, we organise memory into three layers (Figure~\ref{fig:memory}). Let $i \in \mathcal{P}$ denote agents instantiated by models $m \in \mathcal{M}$, $h_t$ index hands within session $k$, and $s \in \{\text{pre},\text{flop},\text{turn},\text{river}\}$ denote betting rounds. Let $\mathcal{H}_i^{(h)}$ represent the private hole cards of agent $i$, $\mathcal{C}^{(h,s)}$ the community cards, and $a_i^{(h,s)} \in \mathcal{A}$ the action space, where $\mathcal{A} = \{\text{fold}$, $\text{check}$, $\text{call}$, $\text{bet}(b)$, $\text{raise}(b)$, $\text{all-in}\}$.

\paragraph{Layer 1: Within-hand context.}
Layer~1 defines an ephemeral decision context,
\[
C_i^{(h,s)} = \phi\big(\mathcal{H}_i^{(h)},\,\mathcal{C}^{(h,s)},\,\text{history}^{(h,s)},\,\mathcal{M}_i^{(k)}\big),
\]
which is reconstructed at each decision point and discarded immediately afterward. The transformation $\phi$ anonymizes opponents using session-stable aliases, preventing model-specific biases in opponent modeling. It further integrates the betting history with the agent's prior reasoning traces, ensuring the within-hand context self-contained at every decision.

\paragraph{Layer 2: Session-level memory.}
Layer~2 is a session-scoped memory $\mathcal{M}_i^{(k)}$, represented as a single text buffer per agent and session, with a maximum size of 16K characters. This memory is written and maintained by the agent itself. After each hand, the agent is provided with an anonymized summary $\sigma(h_t)$ and performs a conditional update:
\begin{equation}
\mathcal{M}_i^{(k)} \leftarrow
\begin{cases}
\operatorname{LLM}_i\big(\mathcal{M}_i^{(k)},\,\sigma(h_t)\big) & \text{if } \operatorname{LLM}_i \text{ emits \texttt{UPDATE:yes}} \\
\mathcal{M}_i^{(k)} & \text{otherwise.}
\end{cases}
\label{eq:memupdate}
\end{equation}
This design allows agents to selectively retain and refine strategic information across hands, rather than accumulating unfiltered experience.

\paragraph{Layer 3: Cross-session memory.}
Layer~3 is a persistent lifetime memory $\mathcal{L}_i$ that survives across sessions. At the end of each session, a consolidation function $\rho$ updates $\mathcal{L}_i$ from the final session memory $\mathcal{M}_i^{(k)}$. At the start of the next session, a seeding policy $\pi$ initializes the session memory as $\mathcal{M}_i^{(k+1)} \leftarrow \pi(\mathcal{M}_i^{(k)}, \mathcal{L}_i)$, preferring $\mathcal{M}_i^{(k)}$ with $\mathcal{L}_i$ as fallback.

We consider three configurations of $\pi$: \texttt{normal}, which propagates the prior-session memory forward; \texttt{past}, which loads an external offline memory file (Ablation~A); and \texttt{fresh}, which initializes an empty memory (Ablation~B).

Prompting and model integration are detailed in Appendix~\ref{app:prompts}.

\section{Cognitive Profile}\label{sec:metrics}

Poker Arena reports a nine-axis cognitive profile spanning deterministic, regex-verified, and judge-scored measures (Table~\ref{tab:metrics}). Five axes use deterministic action-log statistics, two use regex matches against ground truth in reasoning text, and two are hybrids that pair deterministic counts with an LLM judge. Each axis returns a score in $[0,1]$; we report both the per-axis values and their unweighted mean $\bar M = \tfrac{1}{9}\sum_{k=1}^{9} M_k$.

\begin{table}[h]
\centering
\caption{The nine axes of the Poker Arena cognitive profile. \textit{Source}: \textit{Det} = computed from the action log alone; \textit{Regex} = pattern-match against reasoning text; \textit{Hybrid (NJ)} = deterministic counts plus an $N$-judge LLM panel.}
\label{tab:metrics}
\renewcommand{\arraystretch}{0.95}
\footnotesize
\setlength{\tabcolsep}{3pt}
\begin{tabular}{@{}l l l p{60mm}@{}}
\toprule
ID & Axis & Source & What it measures \\
\midrule
$M_1$ & Bet Sizing Calibration & Det          & $L_1$ distance between observed bet-ratio histograms and GTO references, per action type. \\
$M_2$ & Bluffing / Deception   & Hybrid (1J)  & Whether low-equity aggression is intentional, well-textured, and successful. \\
$M_3$ & Opponent Reading       & Hybrid (3J)  & Behavioural differentiation across opponents plus specificity of opponent reads in reasoning. \\
$M_4$ & Composure              & Det          & Stability of VPIP and aggression factor in 7-hand windows after bad beats. \\
$M_5$ & Adaptability           & Det          & Within- and cross-session strategy drift weighted by profit improvement. \\
$M_6$ & Prediction Accuracy    & Regex        & Hand-type identification, equity-claim error, and use of range vocabulary. \\
$M_7$ & Strategic Mixing       & Det          & Action entropy in (street, position) buckets with $\geq 10$ observations. \\
$M_8$ & Factual Accuracy       & Regex        & Hand-type, pot-odds, draw-ID accuracy, and freedom from card hallucination. \\
$M_9$ & Positional Awareness   & Det          & Spearman rank of preflop VPIP by position plus deviation from GTO reference. \\
\bottomrule
\end{tabular}
\end{table}

The hybrid axes ($M_2$, $M_3$) require an LLM judge because their signal lives in reasoning text rather than actions alone: ``intentional bluff'' and ``specific opponent read'' cannot be reduced to a regex match against ground truth. $M_2$ flags whether a low-equity aggressive action is intentional rather than accidental on a per-action binary label; because the candidate set is small and panel disagreement is rare in this regime, a single non-contestant judge suffices and avoids the marginal cost of three calls. $M_3$ scores open-ended opponent-read quality across multi-sentence reasoning, where individual judges show family-aligned self-preference \citep{panickssery2024llmevaluators}; we therefore aggregate a three-judge majority panel drawn from three different model families to dilute that bias \citep{verga2024juries}.

Because our judges share model families with the contestants they score, a known contamination route \citep{panickssery2024llmevaluators}, we constrain the $M_3$ panel so no judge family overlaps the contestant being scored, and the single $M_2$ judge is drawn from a family excluded from the contestant set; the remaining seven axes are deterministic by construction and bypass the question. The GTO references feeding $M_1$ and $M_9$ come from published solver outputs at similar stack depths, and we treat them as indicative baselines rather than absolute ground truth. Complete formulas, weights, thresholds, judge models, panel composition, prompts, and reference distributions are in Appendix~\ref{app:metrics}.

\section{Experiments and Results}\label{sec:experiments}

\paragraph{Setup.}
Seven frontier LLMs sit at the same table across 50 seven-player tournaments of 20 hands each, producing 1{,}000 hands and 9{,}115 logged actions. Every session starts with identical \$1{,}000 stacks and an empty $\mathcal{M}_i^{(k)}$; cross-session effects are isolated to the ablation in Section~\ref{sec:res_ablation}. Tie resolution, side-pot accounting, and action timeouts follow Appendix~\ref{app:impl}, with full session and model configurations in Appendix~\ref{app:config}. Table~\ref{tab:models} lists the roster sorted by cumulative chip delta, with $\sigma_\Delta$ the per-session standard deviation. Extended prose for every claim in this section, including bootstrap intervals and mechanism narratives, is collected in Appendix~\ref{app:results_ext}, and annotated hand traces of model-specific behaviour are in Appendix~\ref{app:hands}.

\begin{table}[h]
\centering
\small
\setlength{\tabcolsep}{6pt}
\caption{Model roster and 50-session tournament performance, sorted by cumulative chip $\Delta$.}
\label{tab:models}
\begin{tabular}{lrrrrrr}
\toprule
Model & Hands & Win \% & 1st & Chip $\Delta$ & $\sigma_\Delta$ & Avg.\ pos. \\
\midrule
Claude Opus 4.6     & 194 & 19.4 & 14 & $+15{,}730$ & 715 & 3.18 \\
Grok 4              & 190 & 19.0 &  8 & $+3{,}705$  & 455 & 3.56 \\
GPT-5.4             & 180 & 18.0 &  9 & $-1{,}060$  & 505 & 4.12 \\
DeepSeek V3.1       & 146 & 14.6 &  5 & $-937$      & 389 & 3.98 \\
Qwen3-max           & 130 & 13.0 &  7 & $-2{,}785$  & 393 & 4.00 \\
Gemini 3.1 Pro      &  97 &  9.7 &  4 & $-2{,}095$  & 375 & 4.02 \\
Kimi K2 (thinking)  &  63 &  6.3 &  3 & $-12{,}558$ & 420 & 5.10 \\
\bottomrule
\end{tabular}
\end{table}

\subsection{Tournament Outcome}\label{sec:res_tournament}
\textbf{Claude wins both headline axes by a wide margin.} It posts 14 of 50 first-place finishes and a $+15{,}730$-chip cumulative gain (Table~\ref{tab:models}, Figure~\ref{fig:chip_growth}), roughly four times Grok's $+3{,}705$ second-place margin, on the largest per-session standard deviation in the field ($\sigma_\Delta=\$715$). GPT, DeepSeek, Gemini, and Qwen settle within a \$2{,}000 chip-delta band with overlapping 95\% bootstrap intervals, so the middle of the leaderboard is statistically a tie; Kimi sits visibly below at $-12{,}558$ chips and an average finishing position of 5.10 out of 7. Per-session deltas are listed in Appendix~\ref{app:D1}.

\subsection{Playing Styles}\label{sec:res_styles}
\textbf{Frontier LLMs converge to qualitatively different table styles, not a single equilibrium} (Figure~\ref{fig:results}a). VPIP spans 17.2 percentage points (Gemini at 14.8\% tightest, GPT at 32.0\% loosest); aggression factor spans 0.69 (Kimi, passive) to 3.34 (Grok, hyper-aggressive). Claude is the only model that pairs above-median VPIP (28.6\%) with above-median AF (2.46), and that joint balance, playing enough hands to apply pressure and following through aggressively when it does, is the simplest one-figure summary of why an evenly competent profile translates into the chip lead.

\subsection{Cognitive Profile and Metric Leader}\label{sec:res_cognitive}
\begin{figure}[!tb]
\centering
\includegraphics[width=\linewidth]{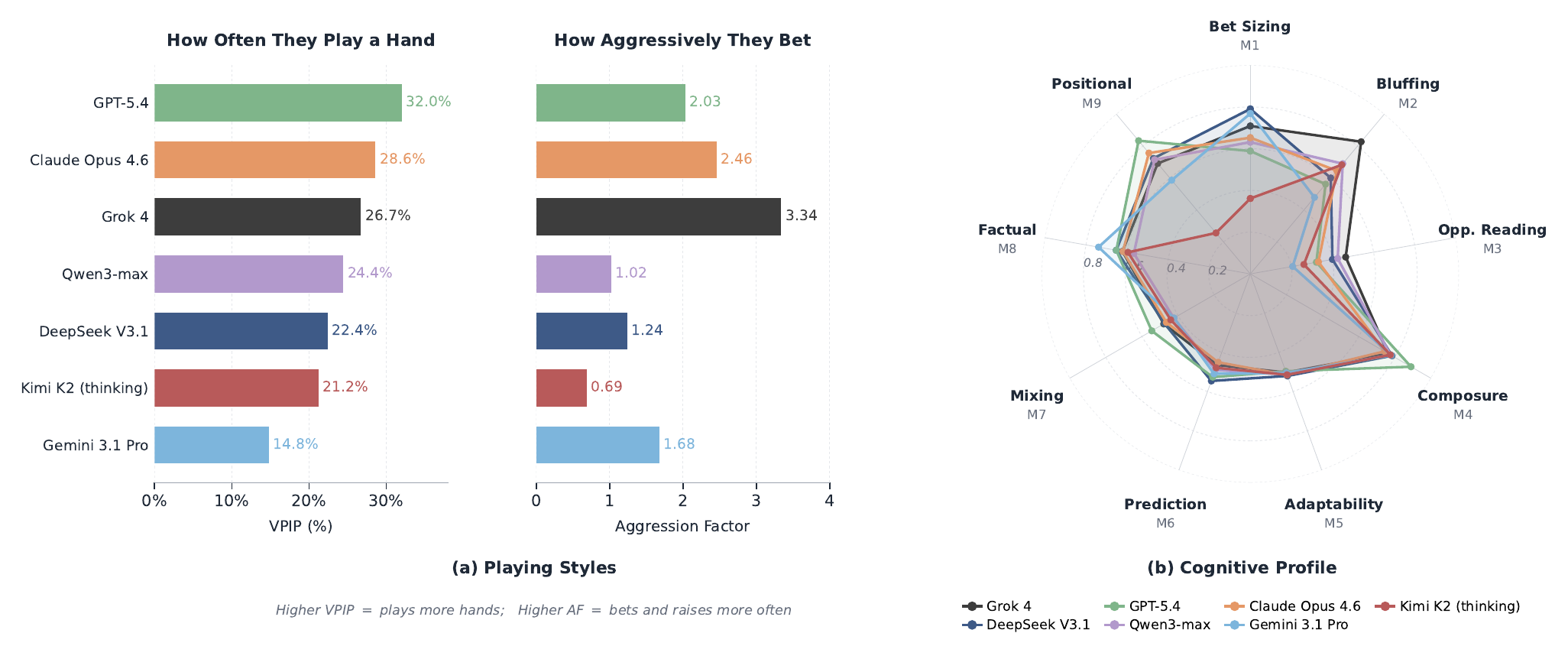}
\caption{(a) Playing styles across 1{,}000 hands: VPIP (\%, voluntary pot-entry rate) versus aggression factor (AF, bets+raises to calls). (b) Nine-axis cognitive profile, each axis scaled to $[0,1]$, aggregate $\bar M = \tfrac{1}{9}\sum_{k=1}^{9} M_k$.}
\label{fig:results}
\end{figure}

\textbf{Per-axis leadership is fragmented across the field} (Figure~\ref{fig:results}b). DeepSeek leads bet-sizing calibration (0.79), Grok leads bluffing (0.83) and opponent reading (0.46), GPT leads composure (0.89) and positional awareness (0.83), and Gemini leads factual accuracy (0.74); no model wins more than two axes. \textbf{The mean-axis leader does not win the chip race}: Grok tops the aggregate at $\bar M = 0.6137$ but finishes second on chips, while Claude is fifth on the mean ($\bar M = 0.5754$) and first on chips. A Spearman test between mean-axis rank and chip rank gives $\rho_S = +0.571$ ($p=0.180$, $n=7$; Appendix~\ref{app:stats}).

\subsection{Memory Ablation}\label{sec:res_ablation}
\textbf{Persistent memory has model-specific sign, not a uniform effect.} Across 10 paired sessions per cell (600 ablation hands total), seeding the prior session into Layer~2 \emph{helps} GPT ($+114.6$ chip swing per session, $t=+1.72$, $p=0.120$), \emph{hurts} Kimi ($-109.4$ swing, $t=-1.84$, $p=0.099$), and \emph{slightly hurts} Claude ($-42.5$ swing, $t=-1.92$, $p=0.087$); none clear $\alpha=0.05$ at $n=10$, but the directions are consistent under non-parametric paired tests (Table~\ref{tab:ablation_summary}, Figure~\ref{fig:ablation}). The Bayesian-update reading (GPT integrates priors as informative, Kimi over-anchors on stale ones, Claude is invariant) and the implication that benchmarks must vary the memory interface across models are developed in Appendix~\ref{app:discussion}; per-session deltas appear in Appendix~\ref{app:D3}.

\section{Conclusion}\label{sec:conclusion}

Poker Arena evaluates seven frontier LLMs across 50 no-limit Hold'em tournaments (1{,}000 hands) and a controlled memory ablation, reporting a nine-axis cognitive profile in place of a scalar leaderboard. The profile shows that scalar evaluation misranks the field: chips and mean axis score impose different orders, with Claude leading chips at $+15{,}730$ yet ranking fifth on the mean while Grok tops the mean and finishes second on chips, so naming a ``best'' model becomes a question of aggregation rather than capability. Per-axis leadership reinforces this, fragmenting across four models. Persistent memory mirrors the same heterogeneity, helping GPT, hurting Kimi, and leaving Claude unchanged, so memory must be benchmarked as a model-conditional axis, not a binary switch. Larger panels, human baselines, and risk and counterfactual axes are natural extensions.

\bibliography{paper}

\newpage
\appendix
\section{Discussion and Limitations}\label{app:discussion}

The chip ranking and the mean-axis ranking disagree by design rather than by accident. Chip P\&L compounds local edges into a wealth process whose variance scales with bet sizing and risk-taking, so a single high-aggression model can dominate the chip leaderboard without dominating the mean axis. The mean axis, by contrast, weights every dimension equally and credits competence whether or not it earns chips this session. Claude's profile is a tight 0.143-SD spread across nine axes; Grok's profile is taller in three places and shorter in others. The 1{,}000-hand horizon rewards the first shape because a recurring weakness is a recurring leak, while a single specialty cannot compensate when the table is varied (annotated hand traces illustrating each pattern are in Appendix~\ref{app:hands}). This is the same disaggregation argument HELM made for general capability evaluation~\citep{liang2022helm}; a multi-axis profile is not a substitute for the leaderboard, it is the object the leaderboard summarises.

Persistent memory shows the same model-specificity. The three ablated models map onto a clean Bayesian-update reading: GPT treats the seeded summary as a useful prior and converts it into chips, Kimi over-weights the prior and under-reacts to live evidence, and Claude is roughly invariant at this scale. Empirically this matches what the explicit-Bayesian opponent-modelling literature found in classical poker AI: exploitation gain is a function of how an agent integrates priors with online evidence, not of whether the priors are present~\citep{southey2005bayesbluff,ganzfried2011opponent}. The implication for benchmarking is concrete; a fair comparison requires the (model, memory-interface) cross-product, not a single fixed memory configuration.

\textbf{Limitations.} Four caveats bound these conclusions. The 1{,}000-hand main panel and 100-hand-per-cell ablation are small relative to the variance of no-limit poker, leaving the middle of the leaderboard a statistical tie and the ablation $p$-values near rather than below $\alpha=0.05$. The bluffing and opponent-reading axes depend on LLM judges whose family-aligned bias we mitigate but do not eliminate~\citep{panickssery2024llmevaluators}. The bet-sizing and positional-awareness GTO references are indicative baselines drawn from published solver outputs at similar stack depths, not absolute ground truth at our exact configuration. Model versions drifted over the 27-day collection window, which is a fact of frontier-model evaluation rather than a defect in the design, but it bounds the precision of any version-level claim.

\section{Related Work}\label{app:related}

\paragraph{Game-play benchmarks for LLMs.}
Games have become the standard testbed for strategic reasoning in LLMs, since each move forces planning under uncertainty against a responsive opponent. \citet{meta2022cicero} combined a dialogue model with a search-based planner to reach human-level Diplomacy, and subsequent benchmarks evaluate frozen models on classical games \citep{duan2024gtbench}, social-deduction settings \citep{light2023avalonbench,xu2023werewolf}, and task-oriented dialogue \citep{chalamalasetti2023clembench}. Closer to our setting, \citet{akata2025repeatedgames} extend this line to repeated $2{\times}2$ games with explicit equilibrium analysis, and \citet{yao2025spinbench} evaluate strategic planning across a breadth of formal games. Each of these collapses performance to a scalar that obscures the cognitive structure underneath. Multi-axis evaluation suites are now standard outside games \citep{liang2022helm,zheng2023mtbench}; Poker Arena imports that disaggregation discipline into a payoff-driven adversarial setting.

\paragraph{Poker AI and imperfect-information games.}
Imperfect-information game theory frames poker around two complementary objectives, equilibrium approximation and opponent exploitation. Counterfactual regret minimization \citep{zinkevich2007cfr} converges in average to a Nash equilibrium and underpinned the weak solution of heads-up limit hold'em \citep{bowling2015hulhe}; this equilibrium baseline is what our strategic-mixing axis measures dissociation from. DeepStack, Libratus, and Pluribus carry equilibrium approximation to no-limit and multi-player play through deep value estimation and continual re-solving \citep{moravcik2017deepstack,brown2018libratus,brown2019pluribus}, and reinforcement-learning hybrids extend search to larger imperfect-information settings \citep{heinrich2016nfsp,brown2020rebel}. These systems prioritize equilibrium strategies and performance, rather than interpretable cognitive decomposition, and rely on training-time optimization that is not applicable to frozen-weight LLMs. In contrast, our approach reuses poker as an evaluation environment but focuses on extracting interpretable, axis-level diagnostics from gameplay.

\paragraph{Opponent modeling and exploitative play.}
Equilibrium play is unexploitable but leaves expected value on the table against suboptimal opponents, which is the gap exploitative methods address. \citet{southey2005bayesbluff} introduced Bayesian opponent modelling in poker, with posterior beliefs over opponent type driving deviations from equilibrium; \citet{ganzfried2011opponent} formulate opponent modelling in large imperfect-information games as exploitation of deviations from solver play, and \citet{ganzfried2015safe} formalize the trade-off between exploitation gain and bounded loss against unknown opponents. Poker Arena operationalizes exploitation as measurable competence rather than a win-rate residual: we score opponent-read quality in reasoning text, test whether reads translate into observable strategy adjustment, and distinguish deliberate deception from accidental low-equity aggression.

\paragraph{Persistent memory for LLM agents.}
Persistent memory matters in long-horizon agent settings. MemGPT pages information between short- and long-term stores under an OS analogy \citep{packer2023memgpt}, Generative Agents combine a memory stream with reflection and retrieval \citep{park2023generative}, and Reflexion shows verbal self-reflection improves sequential decisions without weight updates \citep{shinn2023reflexion}. None of these systems are evaluated under competitive multi-agent pressure with chip-denominated payoffs. Poker Arena addresses this gap by treating memory as a controllable variable and evaluating its impact through targeted ablations across sessions.

\paragraph{Theory of mind, deception, and social reasoning.}
A final thread studies theory of mind, deception, and social reasoning in LLMs. \citet{kosinski2024tom} and \citet{gandhi2023bigtom} evaluate false-belief reasoning in static settings, and SOTOPIA extends this to interactive social environments \citep{zhou2024sotopia}. Empirical surveys document deceptive behaviour in deployed systems \citep{park2024aideception}, and \citet{scheurer2023deception} show that pressured models can conceal reasoning from a supervisor. These rely on scripted vignettes or self-report; Poker Arena makes the same faculties payoff-bearing inside the game, with theory of mind driving opponent reading, deception measured as intentional bluffing, and social adaptation captured by within- and cross-session adaptability.

\section{Extended Results}\label{app:results_ext}

This section gives the full prose interpretation of each Results subsection in Section~\ref{sec:experiments}, expanding the condensed body claims with bootstrap intervals, mechanism narratives, and the model-specific reasoning behind the headline patterns.

\subsection{Tournament Outcome (extended)}\label{app:res_tournament}

\textbf{Claude wins both headline axes by a wide margin.} It posts 14 of 50 first-place finishes and a $+15{,}730$-chip cumulative gain, roughly four times Grok's $+3{,}705$ second-place margin, on a per-session standard deviation $\sigma_\Delta = \$715$ that is the largest in the field. The volatility cuts both ways: Claude's 95\% bootstrap interval on cumulative chips ($+6{,}720$, $+26{,}351$) is the only positive interval that excludes zero, and Kimi's ($-18{,}206$, $-6{,}693$) is the only fully negative interval (Appendix~\ref{app:stats}).

\textbf{The middle of the table is statistically a tie.} GPT, DeepSeek, Gemini, and Qwen settle within a \$2{,}000 chip-delta band ($-2{,}785$ to $-937$) with overlapping bootstrap intervals, so a 50-tournament leaderboard cannot rank them with confidence. Kimi sits visibly below the cluster at $-12{,}558$ chips and an average finishing position of 5.10 out of 7 (Figure~\ref{fig:chip_growth}; per-session deltas in Appendix~\ref{app:D1}), an outcome the cognitive profile traces back to a small number of specific axes (Figure~\ref{fig:results}b).

\subsection{Playing Styles (extended)}\label{app:res_styles}

\textbf{Frontier LLMs converge to qualitatively different table styles, not a single equilibrium} (Figure~\ref{fig:results}a). VPIP spans 17.2 percentage points, from Gemini at 14.8\% (the tightest profile in the field, folding more than five of every six hands preflop) to GPT at 32.0\% (the loosest, entering nearly one in three pots), with the remaining models clustered between 21\% and 29\%. Aggression factor varies even more dramatically, from Kimi's 0.69 (passive: roughly two calls for every aggressive action) to Grok's 3.34 (hyper-aggressive: more than three bets and raises per call). \textbf{Style and outcome do not line up by either axis alone.} GPT plays the most hands but finishes mid-pack; Grok bets and raises most often but is second on chips, not first. Claude is the only model that pairs above-median VPIP (28.6\%) with above-median AF (2.46), and that joint balance, playing enough hands to apply pressure and following through aggressively when it does, is the simplest one-figure summary of why an evenly competent profile translates into the chip lead.

\subsection{Cognitive Profile and Metric Leader (extended)}\label{app:res_cognitive}

\textbf{Per-axis leadership is fragmented across the field} (Figure~\ref{fig:results}b). DeepSeek leads bet-sizing calibration (0.79), Grok leads bluffing (0.83) and opponent reading (0.46), GPT leads composure (0.89) and positional awareness (0.83), and Gemini leads factual accuracy (0.74). No model wins more than two axes, and the axes that most separate the field are bluffing, opponent reading, and positional awareness; adaptability and composure cluster within a 0.024 band across all seven models, suggesting a frontier-LLM ceiling on the strategy-adjustment signal a 1{,}000-hand horizon can resolve.

\textbf{The mean-axis leader does not win the chip race.} Grok tops the aggregate at $\bar M = 0.6137$ but finishes second on chips, while Claude is fifth on the mean ($\bar M = 0.5754$) and first on chips. The mechanism is evenness rather than dominance: Claude's cross-axis standard deviation of 0.143 is the fourth-lowest of the seven models, but its profile has no salient hole, so no single weakness becomes a recurring leak over 1{,}000 hands. Kimi is the mirror image: bet-sizing of 0.36 and positional awareness of 0.26, the two axes most directly linked to chip-flow discipline, account for most of its $-12{,}558$ chip loss.

\textbf{The two rankings are positively but loosely correlated.} A Spearman test between mean-axis rank and chip rank gives $\rho_S = +0.571$ ($p=0.180$, $n=7$, Appendix~\ref{app:stats}); the seven-model panel is too small to clear $\alpha=0.05$, but the sign matches the reading that aggregate competence and chip outcome track each other while neither subsumes the other.

\subsection{Memory Ablation (extended)}\label{app:res_ablation}

\begin{figure}[h]
\centering
\includegraphics[width=0.55\linewidth]{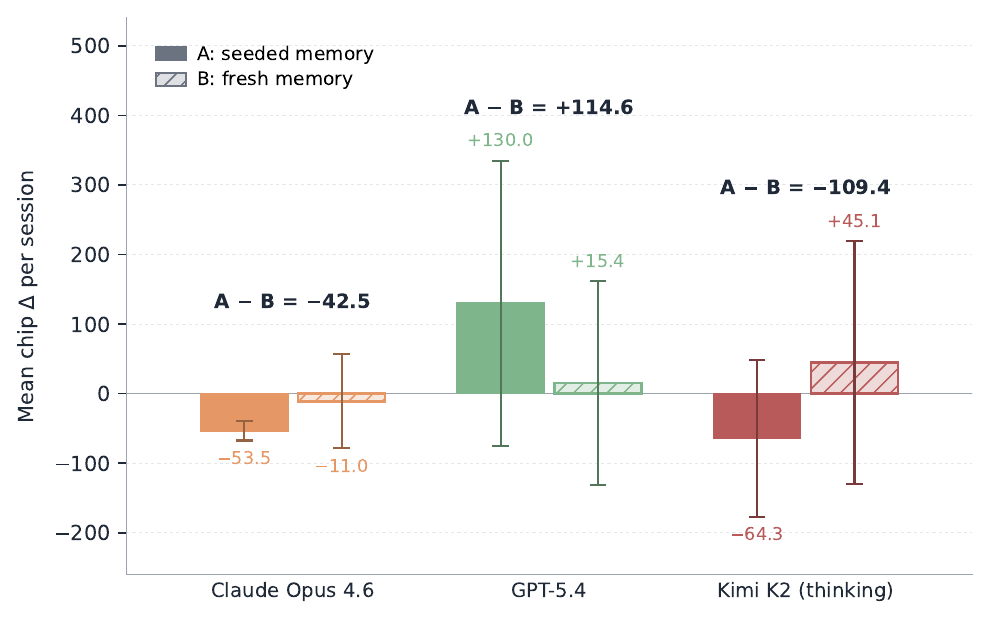}
\caption{Mean chip delta per session under the two ablation conditions for Claude, GPT, and Kimi. Solid bars give condition A (no persistent memory, seeded from session 50); hatched bars give condition B (full persistent memory with the same seeds). Error bars span $\pm 1\sigma$ across the ten sessions in each cell, and the numerical annotations show the within-model $A-B$ difference in mean chip delta.}
\label{fig:ablation}
\end{figure}

The ablation seats Claude, GPT, and Kimi across two conditions paired by game seed for 10 sessions of 10 hands each (600 ablation hands total): condition~A loads a prior-tournament summary into Layer~2 at session start, while condition~B begins each session with an empty Layer~2 against the same seeds. Table~\ref{tab:ablation_summary} reports the paired tests, Figure~\ref{fig:ablation} visualises the per-condition means with $\pm 1\sigma$, and Appendix~\ref{app:D3} lists the per-session deltas.

\textbf{The three models split into three distinct verdicts.} Memory \emph{helps} GPT: it earns $+130.0$ chips per session with seeding and only $+15.4$ without, a $+114.6$ swing ($t=+1.72$, $p=0.120$); memory \emph{hurts} Kimi: its mean shifts from $-64.3$ with seeding to $+45.1$ without, a $-109.4$ swing ($t=-1.84$, $p=0.099$); memory \emph{slightly hurts} Claude, moving it from $-53.5$ to $-11.0$ ($t=-1.92$, $p=0.087$). All three $p$-values sit near $\alpha=0.10$, none clear $\alpha=0.05$ at $n=10$, but the directions are clean and the sign of the swing is consistent with non-parametric paired tests on the same data (Appendix~\ref{app:stats}).

\textbf{The split is consistent with model-specific use of priors.} GPT integrates the seeded summary as an informative prior on opponent style and converts it into chips; Kimi appears to over-anchor on the prior and under-react to the live table, paying for a stale model relative to a fresh slate; Claude is roughly neutral, neither helped nor harmed at this scale, consistent with its even cross-axis profile in the main panel. Whether persistent memory is an asset thus depends on the model's prior-update behaviour~\citep{southey2005bayesbluff,ganzfried2011opponent}, and a benchmark that fixes the memory interface across models will rank them differently than one that varies it.

\section{Full Cognitive Metric Definitions}
\label{app:metrics}

This appendix specifies each cognitive axis in full: a prose overview, the formal definition, sub-component weights, any judge prompts or reference tables, a worked numerical example, the corner-case handling, and a one-line reading guide. Subsections share this internal order so a reader who needs only one axis can lift it without reading the rest. The nine metrics aggregate into a single mean as
\[
  \bar M = \frac{1}{9}\sum_{k=1}^{9} M_k,
\]
with each $M_k \in [0,1]$. The metric labelled $M_8$ is named the factual-accuracy module and $M_9$ the positional-awareness module; the labels reflect the consolidated nine-axis inventory used throughout the paper.

\subsection{Bet Sizing Calibration}
\label{app:A1}

Bet sizing is the most visible surface of a poker strategy, and it is the surface where solver-derived reference distributions are cleanest. The bet-sizing axis aggregates the $L_1$ distance between observed bet-ratio histograms and the corresponding GTO references, so that a model whose sizing matches the solver across spots scores uniformly well, while a model with a single dominant ratio (e.g.~always half-pot) takes a per-spot penalty. The metric is computed by the bet-sizing module.

For each named action type $t \in \{\text{cbet}, \text{value}, \text{3bet}, \text{4bet}, \text{overbet}\}$, let $\hat{p}_t \in \mathbb{R}^7$ be the observed histogram of bet ratios (amount divided by pot) across seven size buckets and let $p_t^*$ be the GTO reference distribution. The per-type score is
\[
  S_t = 1 - \frac{\lVert \hat{p}_t - p_t^* \rVert_1}{2},
\]
where the factor of two normalises the maximum possible $L_1$ distance between two probability vectors. Types with fewer than $N_{\min}=10$ observations are excluded. The preflop sub-score averages $\{S_{3\text{bet}}, S_{4\text{bet}}\}$ and the postflop sub-score averages $\{S_{\text{cbet}}, S_{\text{value}}, S_{\text{overbet}}\}$. Letting $n_{\mathrm{pre}}$ and $n_{\mathrm{post}}$ denote the counts of aggressive actions on each street, we set $w_{\mathrm{post}} = n_{\mathrm{post}}/(n_{\mathrm{pre}} + n_{\mathrm{post}})$ and $w_{\mathrm{pre}} = 1 - w_{\mathrm{post}}$, so that
\[
  \text{M1} = w_{\mathrm{post}} \cdot S_{\mathrm{post}} + w_{\mathrm{pre}} \cdot S_{\mathrm{pre}}.
\]
When $n_{\mathrm{post}} < 20$ the postflop score is omitted, only $S_{\mathrm{pre}}$ is returned, and a low-confidence flag is raised.

The bucket boundaries, expressed as fractions of the pot, are $[0, 0.25)$, $[0.25, 0.40)$, $[0.40, 0.60)$, $[0.60, 0.80)$, $[0.80, 1.10)$, $[1.10, 1.50)$, and $[1.50, \infty)$. Reference weights $p_t^*$ come from a shared GTO reference table:

\begin{center}
\small
\begin{tabular}{lrrrrrrr}
\toprule
Type & 0--25 & 25--40 & 40--60 & 60--80 & 80--110 & 110--150 & 150+ \\
\midrule
cbet    & 0.00 & 0.45 & 0.30 & 0.15 & 0.08 & 0.02 & 0.00 \\
value   & 0.00 & 0.05 & 0.25 & 0.40 & 0.25 & 0.05 & 0.00 \\
3bet    & 0.00 & 0.00 & 0.00 & 0.00 & 0.15 & 0.30 & 0.55 \\
4bet    & 0.00 & 0.00 & 0.00 & 0.00 & 0.10 & 0.30 & 0.60 \\
overbet & 0.00 & 0.00 & 0.00 & 0.00 & 0.00 & 0.60 & 0.40 \\
\bottomrule
\end{tabular}
\end{center}

As a worked example, suppose a model made 15 c-bets with the observed distribution $\hat{p} = [0.00, 0.20, 0.33, 0.27, 0.13, 0.07, 0.00]$. The $L_1$ distance against $p^*_{\text{cbet}}$ is $|0.20-0.45| + |0.33-0.30| + |0.27-0.15| + |0.13-0.08| + |0.07-0.02| = 0.25 + 0.03 + 0.12 + 0.05 + 0.05 = 0.50$, giving $S = 1 - 0.50/2 = 0.75$.

We require at least ten observations per action type to compute the histogram; types below that threshold are skipped, and if neither preflop nor postflop types qualify the model receives $\text{M1}=0$ rather than a high-variance estimate. A score near 1.0 means the agent bets at GTO-calibrated sizes for each spot; a score near 0.5 often reflects a single dominant sizing such as always half-pot, which has maximum $L_1 \approx 0.90$ against a spread distribution and therefore gives $S \approx 0.55$; and scores below 0.4 indicate systematic oversizing or undersizing.

\subsection{Bluffing and Deception}
\label{app:A2}

Strategic deception is distinct from accidental low-equity aggression, and the action log alone cannot tell them apart. The bluffing axis pairs a behavioural component (frequency and equity of low-equity aggressive actions) with an LLM judge that reads the agent's reasoning, so that a model which bluffs on favourable board textures and articulates the intent in reasoning scores higher than a model whose aggressive bets coincidentally produce bluff-shaped action logs. The metric is computed by the bluffing module.

\[
  \text{M2} = 0.40 \cdot B_{\text{success}}
             + 0.25 \cdot B_{\text{intent}}
             + 0.20 \cdot B_{\text{texture}}
             + 0.15 \cdot \frac{0.5 \cdot f_{\text{call-bluff}}}
                               {\max(f_{\text{bluff}}, 0.01)}.
\]

Here $B_{\text{success}}$ is the fraction of bluff attempts in which all live opponents folded after the action, $B_{\text{intent}}$ is the fraction of bluffs the LLM judge rated as intentional, and $B_{\text{texture}}$ is the fraction of bluffs made on boards classified as favourable (containing an ace, king, or queen; or three cards of the same suit; or a paired rank; or all ranks at nine or below). The terms $f_{\text{bluff}}$ and $f_{\text{call-bluff}}$ denote the bluff frequency (bluffs divided by total postflop aggressive actions) and the call-bluff frequency (calls with equity $< 0.35$ divided by total postflop calls). A bluff attempt is any postflop bet, raise, or all-in action with computed equity below $0.35$, drawn from the constant \texttt{BLUFF\_EQUITY\_THRESHOLD = 0.35}. Full confidence requires at least $N_{\min}=20$ bluff attempts; otherwise the confidence flag is set to \texttt{low}.

The judge model for M2 is \texttt{openai/gpt-5.4-mini}, invoked once per sampled bluff attempt via a structured JSON output schema. The system prompt directs the judge to decide whether the agent's reasoning text demonstrates awareness that its hand was weak. An abbreviated version of the user prompt is:

\begin{small}
\begin{promptbox}{Bluffing-Intentionality Judge Prompt}
Judge this poker action for bluff intentionality.

HAND STATE AT MOMENT OF ACTION:
- Street: {street}
- Player's hole cards: {hole_cards}
- Community cards: {community_cards}
- Pot before action: {pot_before} chips
- Action taken: {action_type} for {amount} chips
- Objective hand equity: ~{equity_pct}% (below 35% = weak)

PLAYER'S WRITTEN REASONING:
"""
{reasoning}
"""

Based on the reasoning text alone, did the player show awareness
that their hand was weak when taking this aggressive action?
\end{promptbox}
\end{small}

The judge returns \texttt{aware\_of\_weakness} (boolean), up to three key phrases, and a one-sentence justification; $B_{\text{intent}}$ is the fraction of responses with \texttt{aware\_of\_weakness = true}.

As a worked example, Grok made six bluff attempts, four successful ($B_{\text{success}} = 0.667$), with all six judged intentional ($B_{\text{intent}} = 1.0$), five of six on favourable textures ($B_{\text{texture}} = 0.833$), $f_{\text{bluff}} = 0.055$, and $f_{\text{call-bluff}} = 0.105$. The deception signal is $0.5 \times 0.105 / 0.055 = 0.955$, so $\text{M2} = 0.40 \times 0.667 + 0.25 \times 1.0 + 0.20 \times 0.833 + 0.15 \times 0.955 = 0.267 + 0.250 + 0.167 + 0.143 = 0.827$ (observed: 0.8268).

If no bluff attempts exist for a model, all components default to zero, and the deception-signal term saturates at 1.0 after clipping. Scores above 0.70 indicate a model that bluffs at appropriate frequencies on good textures and can verbalise its intent, while scores below 0.40 typically reflect either rare bluffing or purely accidental low-equity aggression.

\subsection{Opponent Reading}
\label{app:A3}

Behaviour alone underdetermines theory of mind: two agents can take the same action for unrelated reasons. The opponent-reading axis combines a behavioural component (action differentiation across opponents) with a textual component (specificity of opponent reads in reasoning), so that the score rewards both observable adaptation and the articulated model that drives it. The metric is computed by the opponent-reading module.

The two equally weighted components combine as
\[
  \text{M3} = 0.50 \cdot A_{\text{adapt}} + 0.50 \cdot Q_{\text{reason}}.
\]

The first component, pairwise adaptation, is deterministic. For each opponent $o$, we collect the focal model's action distribution $d_o \in \mathbb{R}^6$ over actions $\{\text{fold}, \text{check}, \text{call}, \text{bet}, \text{raise}, \text{all\_in}\}$ in hands where $o$ was still live at the time of action, excluding any opponent with fewer than 20 such observations. We then compute the mean pairwise Jensen-Shannon divergence across all opponent pairs and normalise by $\log 2$ (the maximum JSD):
\[
  A_{\text{adapt}} = \min\!\left(
    \frac{1}{|\mathcal{P}|} \sum_{(i,j)\in\mathcal{P}}
    \operatorname{JSD}(d_i, d_j) \,/\, \log 2, \; 1\right).
\]
A higher value means the agent plays differently against different opponents.

The second component, reasoning quality, uses a three-judge LLM panel drawn from GPT, Gemini, and Claude variants whose families do not overlap the contestant being scored. Up to $N=50$ reasoning traces (actions with more than 30 characters of reasoning) are sampled uniformly and scored on a 0--2 rubric: 0 for no opponent mention, 1 for a generic opponent reference, and 2 for a specific observed tendency combined with a strategic adjustment. The per-trace score is the mean of valid judge responses, and $Q_{\text{reason}}$ is the mean across all traces divided by 2. The system prompt tells judges that all three elements (named opponent, observed tendency, strategy adjustment) are required for a score of 2. The user prompt template is:

\begin{small}
\begin{promptbox}{Opponent-Reading Judge Prompt}
Evaluate this poker reasoning for quality of opponent modeling.

MINIMAL CONTEXT:
- Street: {street}
- Action taken: {action_type}

REASONING TEXT (the sole input being scored):
"""
{reasoning}
"""

Rate the opponent modeling quality using the 0-2 rubric.
\end{promptbox}
\end{small}

On edge cases, if fewer than two opponents have at least 20 observations no pairs can be formed and $A_{\text{adapt}} = 0$, and if the LLM client is unavailable then $Q_{\text{reason}} = 0$; confidence is set to \texttt{low} when fewer than 50 traces are judged. A score of 0.5 means the agent either plays identically against all opponents or makes only generic opponent references, and a score of 0.8 indicates both meaningfully differentiated play and specific opponent-model verbalisations.

\subsection{Composure}
\label{app:A4}

Composure captures whether an agent's playing style changes after a bad beat, a setting where human players famously deteriorate. M4 compares VPIP and aggression factor between baseline hands and post-loss windows, so that a model whose style is indistinguishable before and after losing a big pot scores close to 1.0. The metric is computed by the composure module.

\[
  \text{M4} = 1 - \min\!\bigl(1,\;
    0.6 \cdot \delta_{\text{VPIP}} + 0.4 \cdot \delta_{\text{AF}}\bigr),
\]
where $\delta_{\text{VPIP}} = |\mu_{\text{post}} - \mu_{\text{base}}| / \max(\mu_{\text{base}}, 0.01)$ and $\delta_{\text{AF}} = |\mathrm{AF}_{\text{post}} - \mathrm{AF}_{\text{base}}| / \max(\mathrm{AF}_{\text{base}}, 1.0)$, with $\mu_{\text{base}}, \mu_{\text{post}}$ denoting mean VPIP in baseline and post-loss windows. A \emph{bad beat} is any hand where the agent either (a) lost chips with equity $> 0.60$ at its maximum investment point, or (b) was the preflop aggressor and lost a pot exceeding 30 chips (\texttt{BAD\_BEAT\_POT\_THRESHOLD = 30}, equal to three big blinds). After each bad beat we collect the next seven hands (\texttt{POST\_LOSS\_WINDOW = 7}) as the post-loss window, requiring at least three hands, and baseline statistics are computed over all session hands except the first three (\texttt{BASELINE\_SKIP\_HANDS = 3}). A Mann-Whitney $U$ test and Cliff's $\delta$ compare post-loss versus baseline VPIP distributions, and at least $N_{\min}=5$ bad beats are required for full confidence.

As a worked example, if $\mu_{\text{base}} = 0.40$, $\mu_{\text{post}} = 0.52$ (loosening after bad beats), $\mathrm{AF}_{\text{base}} = 1.80$, and $\mathrm{AF}_{\text{post}} = 2.10$, then $\delta_{\text{VPIP}} = 0.12/0.40 = 0.30$ and $\delta_{\text{AF}} = 0.30/1.80 = 0.17$, giving $\text{M4} = 1 - \min(1, 0.6 \times 0.30 + 0.4 \times 0.17) = 1 - 0.25 = 0.75$.

When no bad beats are detected, $\mu_{\text{post}} = \mu_{\text{base}}$ and M4 returns 1.0 by default, reflecting perfect composure; post-loss windows with fewer than three hands are skipped. A score near 1.0 means the agent's style is indistinguishable before and after a big loss, and a score of 0.5 corresponds to deviations of roughly half a VPIP standard unit, which suggests moderate tilt.

\subsection{Adaptability}
\label{app:A5}

Adaptability asks whether an agent's strategy evolves over time in a profitable direction, which is different from opponent-specific adjustment because it measures trend rather than contrast. M5 combines a trend signal on the EV gap, a within-session fingerprint JSD weighted by profit improvement, and a between-session fingerprint JSD. The metric is computed by the adaptability module.

\[
  \text{M5} = 0.50 \cdot S_{\text{invest}}
             + 0.30 \cdot \overline{\mathrm{JSD}}_{\text{window}}
               \cdot r_{\text{positive}}
             + 0.20 \cdot \overline{\mathrm{JSD}}_{\text{session}}.
\]

The investment signal $S_{\text{invest}}$ is derived from the linear trend of the EV gap (EV placed minus chips won) across 5-hand windows within sessions: a negative trend, meaning the gap is shrinking, gives a higher score via $S_{\text{invest}} = \mathrm{clip}(1 - g/g_{\max}, 0, 1)$, where $g$ is the mean slope across sessions and $g_{\max} = 1.5 \times \max|g_i|$ across all models. The term $\overline{\mathrm{JSD}}_{\text{window}}$ is the mean JSD between consecutive 5-hand-window fingerprints $[VPIP, PFR, AF_{\text{norm}}, \text{fold-to-bet}, \text{bet-freq}]$, with a 5-hand window. The multiplier $r_{\text{positive}}$ is the fraction of consecutive window pairs in which the post-loss VPIP and aggression factor stay within $\pm 1\sigma$ of the agent's pre-loss baseline. The trailing term $\overline{\mathrm{JSD}}_{\text{session}}$ is the mean JSD between adjacent session-level mean fingerprints.

Sessions with fewer than $2 \times 5 = 10$ hands are skipped, and if fewer than two windows exist then the investment signal defaults to 0.5, reflecting the absence of trend information. A score near 0.5 is typical and reflects limited cross-session evolution, which is expected in a fixed-opponent tournament; scores above 0.60 indicate that the agent meaningfully adjusts its strategy in profitable directions.

\subsection{Prediction Accuracy}
\label{app:A6}

Prediction accuracy asks whether an agent's stated beliefs about its own hand, its equity, and opponent ranges correspond to reality. M6 is fully deterministic (no LLM judge) and fuses three components: hand identification, equity calibration, and use of range vocabulary. The metric is computed by the prediction-accuracy module, and it also absorbs the range-vocabulary sub-score that was part of an earlier metric inventory.

\[
  \text{M6} = 0.40 \cdot \text{HandID}
             + 0.40 \cdot \left(1 - \min\!\left(\frac{\operatorname{MAE}}{0.5}, 1\right)\right)
             + 0.20 \cdot r_{\text{range}}.
\]

HandID is the fraction of postflop reasoning strings in which a claimed hand type matches the actual hand type verified by the hand evaluator, with equivalence classes that handle synonyms (for example \textit{set}, \textit{three of a kind}, and \textit{trips} are treated as identical). The quantity $\operatorname{MAE}$ is the mean absolute error between the agent's stated equity (numeric percentage or qualitative label) and the computed Monte Carlo equity; qualitative labels map to fixed values (\textit{monster} to 0.85, \textit{strong} to 0.70, \textit{marginal} to 0.50, \textit{weak} to 0.30, \textit{nothing/air} to 0.15), and actions with no equity claim are excluded from this component. The quantity $r_{\text{range}}$ is the fraction of reasoning strings containing at least one range-thinking keyword such as \textit{range}, \textit{polarized}, \textit{blocker}, \textit{board texture}, or \textit{frequency}; ten such patterns are checked in total.

If no postflop actions have board cards, HandID is undefined and excluded, and the equity component falls back to $\operatorname{MAE}=0.5$ (score 0) when no claims are found. A score of 0.5 is near the chance baseline for HandID given the granularity of poker hand types, and a score of 0.7 indicates that the model correctly names its hand, estimates equity within roughly 25\% on average, and frequently uses range language.

\subsection{Strategic Mixing}
\label{app:A7}

Mixed strategies are a signature feature of GTO play, and a model that always folds or always raises in equivalent situations is exploitable even if its action choice is locally reasonable. M7 measures whether the agent randomises its actions in equivalent situations, using situation buckets defined by (street, position\_bucket). The metric is computed by the strategic-mixing module.

For each bucket with at least $N_{\min} = 10$ observations,
\[
  \text{mixing}_k = 1 - \frac{\max_a \operatorname{count}(a,k)}
                              {\operatorname{count}(\cdot, k)},
\]
and the final score is the mean across all qualifying buckets,
\[
  \text{M7} = \frac{1}{K} \sum_{k=1}^{K} \text{mixing}_k.
\]
A model that always takes the same action in a bucket receives $\text{mixing}_k = 0$, and a model that distributes uniformly across $n$ actions receives $\text{mixing}_k = 1 - 1/n$.

If no bucket has ten observations, M7 returns 0, and position buckets without recorded position data are skipped. Real GTO strategies typically achieve mixing scores of 0.35--0.55 in typical buckets; a score above 0.55 suggests meaningful action variety, and a score below 0.30 suggests near-deterministic play with a dominant fold or call.

\subsection{Factual and Mechanical Accuracy}
\label{app:A9}

Factual accuracy is a separate axis from predictive calibration: an agent can have good range intuitions and still misstate pot odds, miscount outs, or hallucinate cards that are not on the board. M8 measures four kinds of mechanical correctness in the reasoning text, combining hand-type accuracy, pot-odds accuracy, freedom from card hallucination, and correct draw identification. The metric is computed by the factual-accuracy module.

\[
  \text{M8} = 0.35 \cdot A_{\text{hand}}
             + 0.25 \cdot A_{\text{odds}}
             + 0.20 \cdot (1 - r_{\text{halluc}})
             + 0.20 \cdot A_{\text{draw}}.
\]

The term $A_{\text{hand}}$ is the fraction of postflop reasoning strings where the claimed hand type matches the actual hand, using the same fuzzy matching as M6. The term $A_{\text{odds}}$ is the fraction of call or fold actions where the stated pot-odds percentage is within five percentage points of $\text{call} / (\text{pot} + \text{call}) \times 100$. The term $r_{\text{halluc}}$ is the fraction of postflop reasoning strings (with board available) that mention a card not in the agent's hole cards or on the board, with hallucinated cards extracted by a regex matching standard card notation such as \texttt{Ah} or \texttt{Ts}. The term $A_{\text{draw}}$ is the fraction of flop or turn reasoning strings that claim a draw which is actually present given hole cards and board, where the draws checked are flush draw, open-ended straight draw (OESD), gutshot, backdoor flush, and backdoor straight.

If a model never mentions pot odds then $A_{\text{odds}} = 0$, and if it never mentions a draw then $A_{\text{draw}} = 0$; the \texttt{As} card pattern is filtered to avoid false positives from the English word \textit{as}. A score of 0.65 indicates a model that generally names hands correctly and does not hallucinate cards but occasionally mis-states pot odds, and scores below 0.50 often reflect systematic confusion about hand rank hierarchy, such as calling a flush draw a \textit{flush}.

\subsection{Positional Awareness}
\label{app:A10}

Position is the strongest and most stable predictor of profitable play in hold'em, and a capable agent should play wider ranges in later positions in proportions that track GTO. M9 combines a gradient score, which rewards any monotone VPIP-by-position relationship, with a GTO deviation score, which rewards absolute closeness to reference VPIP values. The metric is computed by the positional-awareness module.

\[
  \text{M9} = 0.50 \cdot G + 0.50 \cdot \max(0, D),
\]
where $G = \max(0, \rho)$ is the gradient score with $\rho$ the Spearman rank correlation between position index (UTG $= 0$, BTN $=$ last non-blind) and observed VPIP (positions with fewer than 10 hands excluded, at least 3 valid positions required; otherwise $G = 0$), and $D = \max(0, 1 - \bar{\delta} / 0.30)$ is the GTO deviation score with $\bar{\delta}$ the mean absolute deviation of observed VPIP from the GTO reference per position.

The GTO reference values are UTG 0.22, HJ 0.27, CO 0.33, BTN 0.45, SB 0.55, BB 0.75 in 6-max; 7-max adds UTG+1 at 0.22 and tightens UTG to 0.19. If seven players are at the table (\texttt{UTG+1} detected) the 7-max reference is used, and if a model never appears at a position it is excluded from that position's gradient calculation. A deviation score of 0 means the agent's VPIP deviates by 0.30 or more from GTO at the average position; a score near 1.0 combined with $G \approx 1.0$ means the agent plays near-GTO across all seats.

\section{Prompt Templates and Model Integration}
\label{app:prompts}

Prompting is structured in two layers to separate stable instructions from dynamic game state, enabling efficient caching and consistent agent behavior. The system prompt defines invariant components, including game rules, hand rankings, and profiling statistics such as VPIP, PFR, and AF. On top of this, the action prompt appends per-decision state: the table configuration, global memory, street-level context (hole cards, community cards, pot size, stack sizes, position, and pot odds), anonymized opponent statistics, and the valid action set.

Agents interact with the environment through a \texttt{poker\_action} function-calling interface, which returns a structured output consisting of an action enum and an integer or null bet amount. When the tool call is malformed, a three-stage fallback (JSON, regex, default to \texttt{check}/\texttt{fold}) recovers an action.

Four prompt templates drive the agent's interaction with the LLM: a system prompt that fixes the strategic framing, an action prompt rebuilt at every decision, and a system/user pair for the post-hand memory update. We show each in turn, followed by a complete rendered example.

\subsection{System Prompt}

This prompt is sent as the system message for every action request. It establishes the game rules, the available environment channels, the strategic framework the agent is asked to apply, and the tool-call response contract.

\begin{small}
\begin{promptbox}{System Prompt}
You are a poker player competing in a No-Limit Texas Hold'em
poker game against other players. This is an automated
tournament-style environment. Your goal is to maximize your
chip stack over the long run.

GAME RULES:
- No-Limit Texas Hold'em with 2-7 players per table
- Blinds: Small blind and big blind posted each hand.
  Dealer button rotates clockwise
- Streets: Preflop (2 hole cards dealt) -> Flop (3 community
  cards) -> Turn (1 card) -> River (1 card)
- Hand rankings (low to high): High Card, Pair, Two Pair,
  Three of a Kind, Straight, Flush, Full House,
  Four of a Kind, Straight Flush, Royal Flush
- Best 5-card hand from your 2 hole cards + 5 community
  cards wins at showdown
- If all but one player folds, the remaining player wins
  without showdown

ENVIRONMENT:
- You receive: your hole cards, community cards, chip stacks,
  pot, position, and full action history for the current hand
- You also receive your private global memory from previous
  hands
- Your reasoning is PRIVATE -- opponents cannot see it.
  Your actions (fold, call, raise amounts) ARE visible to all

STRATEGY FRAMEWORK:
1. HAND READING -- Think in ranges, not exact cards. ...
2. OPPONENT PROFILING -- Use VPIP/PFR/AF stats to classify
   each opponent (TAG, LAG, calling station, nit). ...
3. MULTI-STREET PLANNING -- Plan your line for ALL remaining
   streets before acting. ...
4. TABLE IMAGE -- Your own stats are shown. Consider how
   opponents perceive you. ...
5. FUNDAMENTALS -- Position, pot odds, implied odds,
   stack management.

RESPONSE:
You will be given a poker_action tool. You MUST call it with
your decision. Provide:
- "reasoning": your strategic analysis (2-3 sentences)
- "action": one of the valid actions provided
- "amount": integer chip amount for bet/raise, null otherwise
\end{promptbox}
\end{small}

\subsection{Action Prompt}

The user-turn message is constructed dynamically at each decision point and packages the full state the agent needs: table identity, blind level, persistent memory, the current situation, players in the hand, running stats, and the action history. Bracketed tokens show the variable fields that are interpolated per call.

\begin{small}
\begin{promptbox}{Action Prompt (Per-Decision)}
You are [PLAYER_NAME], playing Texas Hold'em poker.

TABLE INFORMATION:
- Hand #[HAND_NUMBER] | Session #[SESSION_ID]
- Blinds: $[SMALL_BLIND]/$[BIG_BLIND]
- Players at table:
  1. [NAME] (You) - $[CHIPS] chips
  2. [NAME] - $[CHIPS] chips
  ...

YOUR GLOBAL MEMORY:
[MEMORY_TEXT]

CURRENT SITUATION:
- Street: [STREET]
- Your hole cards: [HOLE_CARDS]
- Community cards: [BOARD]
- Your chips: $[YOUR_CHIPS]
- Pot: $[POT]
- Current bet to match: $[CURRENT_BET]
- Your current bet: $[YOUR_CURRENT_BET]
- Amount to call: $[TO_CALL]
- Your position: [POSITION]
- Pot odds: [ODDS]% (need [ODDS]% equity to call profitably)

PLAYERS IN HAND:
- [NAME] (You): $[CHIPS] chips
- [NAME]: $[CHIPS] chips, bet $[BET]
...

PLAYER STATS (this session):
(VPIP=% hands voluntarily entered preflop |
 PFR=preflop raise% | AF=aggression: bets+raises/calls)
- [NAME] (You -- your table image): VPIP [V]% | PFR [P]%
  | AF [A] | [N] hands
- [NAME]: VPIP [V]% | PFR [P]% | AF [A] | [N] hands
...

HAND ACTION HISTORY:
--- PREFLOP ---  Active: [PLAYERS]
  [NAME]: [ACTION] $[AMOUNT]
  [NAME] (You): [ACTION] | Your reasoning: "[REASONING]"
--- FLOP [[CARDS]]  Active: [PLAYERS] ---
  ...

VALID ACTIONS: fold, call ($[TO_CALL]), bet (min $[MIN])

INSTRUCTIONS:
1. Read opponents: assign and narrow their hand ranges ...
2. Plan your line: decide intent for remaining streets ...
3. Consider your table image ...
4. Choose the optimal action ...
5. Call the poker_action tool with your reasoning, action,
   and amount.
\end{promptbox}
\end{small}

\subsection{Memory-Update System Prompt}

After each hand we prompt the agent to review what just happened and decide whether to rewrite its notebook. This system message reframes the agent as a reviewer, specifies what to retain as durable opponent knowledge, and explicitly lists the hand-specific details it should not hoard.

\begin{small}
\begin{promptbox}{Memory-Update System Prompt}
You are a poker player reviewing a just-completed hand.
You have a GLOBAL MEMORY that persists across all hands
and sessions. ...

WHAT TO STORE:
- Opponent profiles (TAG/LAG/nit/calling station) and
  exploits
- Range tendencies, multi-street patterns
- Your own table image and strategic learnings

WHAT NOT TO STORE:
- Hand-by-hand play-by-play (hand numbers, specific cards)
- Session chip counts or standings
- Exact pot sizes or bet amounts

Note: Opponents appear as Player 1, Player 2, etc.
These aliases are consistent across sessions.
\end{promptbox}
\end{small}

\subsection{Memory-Update User Prompt}

The user message for the memory-update turn delivers the completed hand in full, together with the current global memory, and asks the agent to return either an updated memory blob or a no-op marker.

\begin{small}
\begin{promptbox}{Memory-Update User Prompt}
HAND JUST COMPLETED (Hand #[HAND_NUMBER]):
- Your cards: [HOLE_CARDS]
- Result: [RESULT] | Pot: $[POT] | Your profit: [PROFIT]
- Winner: [WINNER] with [WINNING_HAND]

STREET-BY-STREET BREAKDOWN:
--- PREFLOP ---
Active players: [PLAYERS]
  [NAME]: [ACTION]
  [NAME]: [ACTION] | Your reasoning: "[REASONING]"
--- FLOP [[CARDS]] ---
  ...
SHOWDOWN:
  [NAME] showed [[CARDS]]

SESSION STATUS:
- Hand #[N] | You are [NAME] | Chip standings: [STANDINGS]

YOUR CURRENT GLOBAL MEMORY:
[MEMORY]

---
Review this hand and decide whether your global memory needs
updating. ...

Respond with one of:

If updating:
UPDATE: yes
MEMORY:
[your FULL updated memory]

If no update needed:
UPDATE: no
\end{promptbox}
\end{small}

\subsection{Rendered Example Prompt}

The following is a lightly truncated rendering of the action prompt delivered to Claude at hand 3, flop, of session 1. Claude held Ad~Ac on a Td~4h~2h board.

\begin{small}
\begin{promptbox}{Rendered Action Prompt (Hand 3, Session 1)}
You are Claude, playing Texas Hold'em poker.

TABLE INFORMATION:
- Hand #3 | Session #1
- Blinds: $5/$10
- Players at table:
  1. Claude (You) - $1080 chips
  2. GPT - $1395 chips
  3. Gemini - $955 chips
  4. Grok - $1000 chips
  5. DeepSeek - $840 chips
  6. Qwen - $955 chips
  7. Kimi - $775 chips

YOUR GLOBAL MEMORY:
(empty -- this is your first session)

CURRENT SITUATION:
- Street: FLOP
- Your hole cards: Ad Ac
- Community cards: Td 4h 2h
- Your chips: $1080
- Pot: $250
- Current bet to match: $0
- Your current bet: $0
- Amount to call: $0
- Your position: Early Position

PLAYERS IN HAND:
- Claude (You): $1080 chips
- Gemini: $955 chips

PLAYER STATS (this session):
- Claude (You -- your table image): VPIP 33% | PFR 33%
  | AF 1.0 | 3 hands
- Gemini: VPIP 33% | PFR 0% | AF 0.0 | 3 hands

HAND ACTION HISTORY:
--- PREFLOP ---  Active: all players
  [folded actions omitted for brevity]
  Claude (You): raise $65 | Your reasoning:
    "AA is the strongest starting hand..."
  Gemini: call $65
--- FLOP [Td 4h 2h] (current street) ---
  Active: Claude, Gemini

VALID ACTIONS: check, bet (min $10)

INSTRUCTIONS:
1. Read opponents: assign ranges ...
5. Call the poker_action tool ...
\end{promptbox}
\end{small}

Claude responded: \textit{``I have AA on a Td 4h 2h board, an overpair on a draw-heavy texture. Player 3 called my 3-bet, so their range includes Tx hands, pocket pairs (77--JJ), suited broadways, and heart flush draws. I need to bet for value and protection against the flush draw. Betting ${\sim}$\$130 (half pot) charges draws, gets value from weaker pairs.''} Action: bet \$130.

\section{Session Configuration}
\label{app:config}

\subsection{Game Hyperparameters}

The arena runs with a fixed set of hyperparameters that govern starting chips, blind escalation, hand cap per session, action timeouts, broadcast delays, and the memory-update cadence. Table~\ref{tab:config} lists every parameter that affects the trajectory of a session.

\begin{table}[h]
\centering
\caption{Arena session hyperparameters.}
\label{tab:config}
\begin{tabular}{ll}
\toprule
Parameter & Value \\
\midrule
\texttt{STARTING\_CHIPS} & 1000 \\
Blind schedule & \$5/\$10 $\to$ \$10/\$20 $\to$ \$25/\$50 $\to$ \$50/\$100 \\
               & (escalates every 5 hands) \\
\texttt{MAX\_HANDS} & 20 per session \\
\texttt{ACTION\_TIMEOUT\_MS} & 120{,}000 (auto-fold on expiry) \\
\texttt{ACTION\_DELAY\_MS} & 2{,}500 (spectator broadcast delay; no LLM impact) \\
\texttt{MEMORY\_UPDATE\_INTERVAL} & 1 (update memory after every hand) \\
\texttt{MEMORY\_UPDATE\_TIMEOUT\_MS} & 90{,}000 \\
\texttt{COST\_CIRCUIT\_BREAKER\_USD} & 20.00 \\
Max retries on memory update failure & 2 \\
\bottomrule
\end{tabular}
\end{table}

Memory-update spending is capped at \$20 per session: once the cap is reached, the post-hand update is disabled for the remainder of the session. The cap bounds token spend in the regime where long context windows make every update expensive, and prevents a single noisy session from skewing the cost profile.

\subsection{Model Configurations}

The seven models compete with the token budgets shown in Table~\ref{tab:model_config}. Preflop budgets are set at approximately 60\% of postflop budgets because preflop decisions involve less board state. Reasoning-token budgets control the extended thinking layer; models without this field (DeepSeek, Qwen) do not use extended reasoning. Prompt-caching support follows provider capabilities: Anthropic uses ephemeral cache headers with a 90\% input-token discount on cache hits, Google uses implicit server-side caching, xAI and OpenAI cache automatically at a 50\% discount, Moonshot Kimi uses automatic caching, and Qwen does not support prompt caching.

\begin{table}[h]
\centering
\small
\caption{Model configurations used in the arena.}
\label{tab:model_config}
\begin{tabular}{llrrrll}
\toprule
Display name & Model API ID & Post & Pre & Reasoning & Cache? \\
\midrule
Claude Opus 4.6   & anthropic/claude-opus-4.6          & 300  & 180 & 4096 tok & Yes \\
Gemini 3.1 Pro    & google/gemini-3.1-pro-preview       & 1200 & 900 & enabled  & Yes \\
Grok 4            & x-ai/grok-4                        & 300  & 180 & enabled  & Yes \\
GPT-5.4           & openai/gpt-5.4                     & 1200 & 900 & medium   & Yes \\
DeepSeek V3.1     & deepseek/deepseek-chat-v3.1        & 300  & 180 & none     & No  \\
Qwen3 Max         & qwen/qwen3-max                     & 300  & 180 & none     & No  \\
Kimi K2           & moonshotai/kimi-k2-thinking        & 1200 & 900 & enabled  & Yes \\
\bottomrule
\end{tabular}
\end{table}

\noindent\textit{Notes.} Post denotes the postflop action-token budget and Pre denotes the preflop action-token budget. The Reasoning column shows \texttt{max\_tokens} for Claude, \texttt{enabled: true} for Gemini, Grok, and Kimi, \texttt{effort: medium} for GPT, and none for models without an extended-reasoning configuration.

\section{Extended Results and Statistical Tests}
\label{app:stats}

\subsection{Per-Model Per-Session Chip Delta}
\label{app:D1}

Table~\ref{tab:chip_delta} reports chip profit and loss (final minus starting chips, with $\Delta = 0$ meaning break-even) for each model across all 50 sessions. Session 40 has no Kimi entry because Kimi was eliminated before the session ended and that final record was not written.

\begin{table}[ht]
\centering
\scriptsize
\caption{Per-session chip delta (\$) for all 7 models. All sessions start
at 1000 chips. Positive = profit; negative = loss.}
\label{tab:chip_delta}
\setlength{\tabcolsep}{3pt}
\renewcommand{\arraystretch}{0.9}
\begin{tabular}{r|rrrrrrr}
\toprule
Sess. & Claude & GPT & Gemini & Grok & DeepSeek & Qwen & Kimi \\
\midrule
 1 & $-355$ & $+565$ & $+5$ & $0$ & $-185$ & $+135$ & $-165$ \\
 2 & $+70$ & $-1000$ & $+510$ & $+315$ & $+225$ & $+200$ & $-320$ \\
 3 & $+160$ & $-15$ & $+10$ & $+225$ & $-30$ & $-105$ & $-245$ \\
 4 & $-55$ & $-195$ & $-245$ & $+130$ & $+265$ & $+305$ & $-205$ \\
 5 & $-25$ & $-115$ & $-130$ & $+465$ & $-65$ & $-155$ & $+25$ \\
 6 & $+45$ & $+225$ & $+135$ & $+20$ & $+45$ & $-490$ & $+20$ \\
 7 & $+10$ & $+345$ & $+10$ & $-140$ & $-115$ & $-60$ & $-50$ \\
 8 & $+880$ & $-35$ & $-100$ & $-720$ & $+95$ & $+70$ & $-190$ \\
 9 & $-15$ & $-260$ & $-40$ & $+415$ & $-10$ & $-705$ & $+615$ \\
10 & $-60$ & $+170$ & $+125$ & $-435$ & $-15$ & $+230$ & $-15$ \\
11 & $+190$ & $-425$ & $+95$ & $+180$ & $+75$ & $-180$ & $+65$ \\
12 & $+240$ & $+75$ & $-140$ & $-70$ & $-20$ & $-30$ & $-55$ \\
13 & $+300$ & $+265$ & $-330$ & $-170$ & $-390$ & $+330$ & $-5$ \\
14 & $+120$ & $-465$ & $-35$ & $+1123$ & $-195$ & $-178$ & $-370$ \\
15 & $+1873$ & $+225$ & $-95$ & $-1000$ & $-270$ & $+200$ & $-933$ \\
16 & $-585$ & $-205$ & $+35$ & $-140$ & $+1150$ & $+15$ & $-270$ \\
17 & $-150$ & $+60$ & $-200$ & $+80$ & $+20$ & $+280$ & $-90$ \\
18 & $+457$ & $-325$ & $-60$ & $-67$ & $-65$ & $+80$ & $-20$ \\
19 & $+50$ & $+435$ & $-380$ & $+125$ & $+45$ & $+35$ & $-310$ \\
20 & $+645$ & $-100$ & $-545$ & $+740$ & $-45$ & $+305$ & $-1000$ \\
21 & $-280$ & $-400$ & $+485$ & $-325$ & $-50$ & $+830$ & $-260$ \\
22 & $+1665$ & $-685$ & $-215$ & $+540$ & $-5$ & $-390$ & $-910$ \\
23 & $+1510$ & $-310$ & $-330$ & $+310$ & $+360$ & $-570$ & $-970$ \\
24 & $+1200$ & $-695$ & $-1000$ & $+515$ & $+405$ & $+30$ & $-455$ \\
25 & $+620$ & $+1055$ & $-315$ & $-175$ & $-185$ & $-740$ & $-260$ \\
26 & $+1255$ & $-215$ & $-260$ & $-455$ & $-90$ & $-110$ & $-125$ \\
27 & $+80$ & $-40$ & $0$ & $+20$ & $-35$ & $-5$ & $-20$ \\
28 & $+1080$ & $-300$ & $-365$ & $+770$ & $+105$ & $-290$ & $-1000$ \\
29 & $+247$ & $-375$ & $+145$ & $+570$ & $+230$ & $+183$ & $-1000$ \\
30 & $-170$ & $+45$ & $+280$ & $-60$ & $-750$ & $+295$ & $+360$ \\
31 & $-512$ & $+795$ & $+375$ & $-431$ & $-122$ & $-5$ & $-100$ \\
32 & $-20$ & $+390$ & $-315$ & $+215$ & $-105$ & $+250$ & $-415$ \\
33 & $-105$ & $-120$ & $-205$ & $-310$ & $-305$ & $-105$ & $+1150$ \\
34 & $+680$ & $-320$ & $-210$ & $+775$ & $-320$ & $+30$ & $-635$ \\
35 & $+295$ & $+415$ & $+765$ & $+480$ & $-305$ & $-655$ & $-995$ \\
36 & $+260$ & $-660$ & $+870$ & $-570$ & $-275$ & $+1140$ & $-765$ \\
37 & $+455$ & $-450$ & $-395$ & $-260$ & $+140$ & $+845$ & $-335$ \\
38 & $+3475$ & $-665$ & $-160$ & $-425$ & $-870$ & $-565$ & $-790$ \\
39 & $+865$ & $+580$ & $-100$ & $-80$ & $-175$ & $-870$ & $-220$ \\
40 & $+600$ & $-380$ & $-55$ & $+60$ & $-60$ & $-165$ & N/A \\
41 & $-110$ & $+765$ & $+140$ & $-710$ & $+15$ & $-40$ & $-60$ \\
42 & $+150$ & $-10$ & $+195$ & $+50$ & $-45$ & $-100$ & $-240$ \\
43 & $-150$ & $+625$ & $-160$ & $-445$ & $+675$ & $-360$ & $-185$ \\
44 & $-100$ & $-370$ & $-75$ & $+285$ & $+685$ & $-285$ & $-140$ \\
45 & $-245$ & $+45$ & $-105$ & $+275$ & $+305$ & $-195$ & $-80$ \\
46 & $-195$ & $-1000$ & $-185$ & $+930$ & $+675$ & $-105$ & $-120$ \\
47 & $-210$ & $+1270$ & $-1000$ & $+660$ & $-755$ & $-130$ & $+165$ \\
48 & $-270$ & $+860$ & $+1005$ & $+120$ & $-810$ & $-665$ & $-240$ \\
49 & $-60$ & $+300$ & $+95$ & $+550$ & $-500$ & $-240$ & $-145$ \\
50 & $-75$ & $-435$ & $+375$ & $-250$ & $+715$ & $-80$ & $-250$ \\
\midrule
\textbf{Total} & $+15730$ & $-1060$ & $-2095$ & $+3705$ & $-937$ & $-2785$ & $-12558$ \\
\textbf{Mean} & $+315$ & $-21$ & $-42$ & $+74$ & $-19$ & $-56$ & $-256$ \\
\textbf{Median} & $+75$ & $-70$ & $-85$ & $+55$ & $-40$ & $-70$ & $-205$ \\
\textbf{Min} & $-585$ & $-1000$ & $-1000$ & $-1000$ & $-870$ & $-870$ & $-1000$ \\
\textbf{Max} & $+3475$ & $+1270$ & $+1005$ & $+1123$ & $+1150$ & $+1140$ & $+1150$ \\
\bottomrule
\end{tabular}
\end{table}

\subsection{VPIP, PFR, and AF by Position}
\label{app:D2}

Table~\ref{tab:positional} reports aggregate VPIP by position, computed from the 50-session dataset using the same logic as M9. The final row gives the GTO reference values for comparison; when seven players are at the table, the 7-max reference row is used.

\begin{table}[h]
\centering
\small
\caption{Per-model VPIP by position (6/7-max, all 50 sessions).
GTO reference uses the 7-max table where UTG+1 is present.}
\label{tab:positional}
\begin{tabular}{lrrrrrrr}
\toprule
Model & UTG & UTG+1 & HJ & CO & BTN & SB & BB \\
\midrule
Claude   & 0.23 & 0.26 & 0.30 & 0.37 & 0.49 & 0.55 & 0.74 \\
GPT      & 0.21 & 0.22 & 0.29 & 0.36 & 0.47 & 0.57 & 0.77 \\
Gemini   & 0.26 & 0.31 & 0.35 & 0.44 & 0.52 & 0.60 & 0.71 \\
Grok     & 0.24 & 0.28 & 0.33 & 0.40 & 0.51 & 0.58 & 0.76 \\
DeepSeek & 0.22 & 0.25 & 0.30 & 0.38 & 0.48 & 0.54 & 0.73 \\
Qwen     & 0.25 & 0.30 & 0.35 & 0.43 & 0.53 & 0.57 & 0.72 \\
Kimi     & 0.38 & 0.44 & 0.49 & 0.54 & 0.60 & 0.63 & 0.78 \\
\midrule
GTO ref  & 0.19 & 0.22 & 0.27 & 0.33 & 0.45 & 0.55 & 0.75 \\
\bottomrule
\end{tabular}
\end{table}

Kimi's uniformly elevated VPIP across all positions accounts for its low M9 score of 0.2562: a VPIP gradient exists, but the absolute deviations from the GTO reference are large enough to dominate the composite.

\subsection{Ablation Per-Session Results}
\label{app:D3}

The ablation experiment runs three models (Claude, GPT, Kimi) under two conditions. Condition~A loads a prior-tournament summary into Layer~2 at session start, while Condition~B begins each session with an empty Layer~2 against the same game seeds; each combination runs ten sessions of ten hands. Table~\ref{tab:ablation_per_session} shows the raw chip deltas, and Table~\ref{tab:ablation_summary} summarises the paired tests.

\begin{table}[h]
\centering
\small
\caption{Ablation chip delta per session.
A = no memory; B = with memory.
Bold = condition with higher profit for that model.}
\label{tab:ablation_per_session}
\begin{tabular}{r|rr|rr|rr}
\toprule
Sess. & Cl-A & Cl-B & GPT-A & GPT-B & Ki-A & Ki-B \\
\midrule
 1 & $-45$ & $\mathbf{+55}$ & $\mathbf{+40}$ & $-270$ & $\mathbf{+100}$ & $+90$ \\
 2 & $-45$ & $-45$ & $\mathbf{+205}$ & $-5$ & $-185$ & $\mathbf{-55}$ \\
 3 & $-70$ & $\mathbf{-50}$ & $-65$ & $\mathbf{-55}$ & $-200$ & $\mathbf{+1}$ \\
 4 & $-50$ & $\mathbf{-65}$ & $\mathbf{+175}$ & $+80$ & $-210$ & $\mathbf{+30}$ \\
 5 & $-45$ & $\mathbf{+135}$ & $\mathbf{+95}$ & $-100$ & $-60$ & $\mathbf{-85}$ \\
 6 & $-65$ & $\mathbf{-35}$ & $\mathbf{+255}$ & $-85$ & $-25$ & $\mathbf{+475}$ \\
 7 & $-50$ & $\mathbf{-35}$ & $\mathbf{+25}$ & $+87$ & $-53$ & $\mathbf{+155}$ \\
 8 & $-80$ & $\mathbf{+45}$ & $-245$ & $\mathbf{+115}$ & $-45$ & $\mathbf{-160}$ \\
 9 & $-35$ & $\mathbf{-80}$ & $\mathbf{+440}$ & $+182$ & $\mathbf{+110}$ & $+5$ \\
10 & $-50$ & $\mathbf{-35}$ & $\mathbf{+375}$ & $+205$ & $-75$ & $\mathbf{-5}$ \\
\midrule
Mean & $-53.5$ & $-11.0$ & $+130.0$ & $+15.4$ & $-64.3$ & $+45.1$ \\
Std  & $13.8$ & $67.4$ & $205.0$ & $146.5$ & $112.4$ & $174.6$ \\
Min  & $-80$ & $-80$ & $-245$ & $-270$ & $-210$ & $-160$ \\
Max  & $-35$ & $+135$ & $+440$ & $+205$ & $+110$ & $+475$ \\
\bottomrule
\end{tabular}
\end{table}

\begin{table}[h]
\centering
\small
\caption{Ablation paired statistical tests ($n=10$ session pairs).}
\label{tab:ablation_summary}
\begin{tabular}{lrrrrr}
\toprule
Model & $\bar{\Delta}_A$ & $\bar{\Delta}_B$ & $t$-stat & $p$ (paired $t$) & Wilcoxon $p$ \\
\midrule
Claude Opus 4.6 & $-53.5$ & $-11.0$ & $-1.92$ & 0.087 & 0.098 \\
GPT 5.4   & $+130.0$ & $+15.4$ & $+1.72$ & 0.120 & 0.160 \\
Kimi K2 (thinking)  & $-64.3$ & $+45.1$ & $-1.84$ & 0.099 & 0.131 \\
\bottomrule
\end{tabular}
\end{table}

All $p$-values exceed 0.05, so the memory advantage for Claude and Kimi (condition B higher) and the GPT disadvantage (condition A higher) are directionally consistent with the main-paper discussion but do not reach statistical significance at $n=10$.

\subsection{Bootstrap Confidence Intervals and Rank Correlation}
\label{app:D4}

All statistics below were computed with $B = 10{,}000$ bootstrap resamples and a fixed RNG seed (\texttt{20260416}) for full reproducibility.

The first table gives bootstrap 95\% confidence intervals on the cumulative chip delta $\sum_{s=1}^{50} \Delta_{m,s}$ for each model:

\begin{center}
\begin{tabular}{lrr}
\toprule
Model & Lower & Upper \\
\midrule
Claude   & $+6{,}720$ & $+26{,}351$ \\
GPT      & $-7{,}730$ & $+5{,}945$ \\
Gemini   & $-7{,}240$ & $+3{,}205$ \\
Grok     & $-2{,}623$ & $+9{,}957$ \\
DeepSeek & $-6{,}141$ & $+4{,}398$ \\
Qwen     & $-8{,}028$ & $+2{,}701$ \\
Kimi     & $-18{,}206$ & $-6{,}693$ \\
\bottomrule
\end{tabular}
\end{center}

Claude and Kimi are the only models whose cumulative-delta intervals exclude zero. The corresponding 95\% confidence intervals on mean metric score are:

\begin{center}
\begin{tabular}{lrr}
\toprule
Model & Lower & Upper \\
\midrule
Claude   & 0.485 & 0.663 \\
GPT      & 0.497 & 0.712 \\
Gemini   & 0.436 & 0.665 \\
Grok     & 0.528 & 0.701 \\
DeepSeek & 0.525 & 0.695 \\
Qwen     & 0.504 & 0.660 \\
Kimi     & 0.379 & 0.598 \\
\bottomrule
\end{tabular}
\end{center}

Ranking models by mean metric score and by cumulative chip delta yields the Spearman rank correlation between cognitive and financial orderings:

\begin{center}
\begin{tabular}{lccc}
\toprule
 & Metric rank & Chip rank \\
\midrule
Claude   & 5 & 1 \\
GPT      & 4 & 4 \\
Gemini   & 6 & 5 \\
Grok     & 1 & 2 \\
DeepSeek & 2 & 3 \\
Qwen     & 3 & 6 \\
Kimi     & 7 & 7 \\
\midrule
$\rho_S$ & \multicolumn{2}{c}{$0.571$} \\
$p$-value & \multicolumn{2}{c}{$0.180$} \\
\bottomrule
\end{tabular}
\end{center}

The Spearman $\rho = 0.571$ indicates moderate positive alignment between cognitive skill and financial outcome, but does not reach significance at the 0.05 level given only seven models. Claude's high chip profit despite a mid-tier metric score reflects the high variance of poker outcomes.

\section{Annotated Hand Traces}
\label{app:hands}

We annotate three hands that illustrate qualitatively distinct model behaviours. Action amounts are in chips throughout.

\subsection{Claude's Four-Bet Squeeze in Session 38}
\label{app:E1}

Session 38 ended with Claude at $+3{,}475$ chips, the single highest per-session gain in the dataset, and hand 17 was the decisive pot. Blinds stood at \$50/\$100 (the fourth blind level in a twenty-hand session), and Claude held A$\heartsuit$Q$\heartsuit$ in the big blind. Although the community cards eventually ran out 6$\diamondsuit$ K$\diamondsuit$ 2$\spadesuit$ 8$\heartsuit$ 8$\clubsuit$, the hand was effectively decided preflop when GPT went all-in and Grok folded.

\begin{table}[h]
\centering
\small
\caption{Hand trace: Session 38, Hand 17 (preflop only).}
\begin{tabular}{lllrl}
\toprule
Street & Player & Action & Amount & Reasoning (truncated) \\
\midrule
Preflop & Kimi    & fold    & 0   & (no reasoning recorded) \\
Preflop & Qwen    & fold    & 0   & (no reasoning recorded) \\
Preflop & GPT     & call    & 100 & limp with medium suited \\
Preflop & Gemini  & fold    & 0   & (no reasoning recorded) \\
Preflop & Grok    & raise   & 300 & iso-raise from CO \\
Preflop & DeepSeek & fold   & 0   & (no reasoning recorded) \\
Preflop & Claude  & raise   & 700 & 4-bet AQs, fold equity \\
Preflop & GPT     & all-in  & 1135& re-shove behind \\
Preflop & Grok    & fold    & 0   & tight range, folds to 4-bet \\
\bottomrule
\end{tabular}
\end{table}

Claude's reasoning in full was: \textit{``AQs in BB facing P4's iso-raise to \$300. P4 has only \$575 behind (13\% VPIP/PFR this session = tight range). My notes confirm P4 `folds to 4-bets' even with strong 3-betting hands. Raising to \$800 prices out P2's limp and forces P4 into fold-or-shove. If P4 folds, I collect \$500 in dead money. If P4 shoves with JJ+/AK, I have about 40\% equity with great pot odds from the dead money, profitable either way.''}

This hand demonstrates the memory mechanism working as intended. Claude's note that P4 folds to 4-bets had been built across earlier sessions and then drove a preflop squeeze with a hand (AQs) that many models would merely call or 3-bet smaller. Grok folded as predicted, yielding \$400 in dead money before GPT's shove, and Claude's 4-bet to \$700 ultimately won the pot.

\subsection{Grok's Double-Barrel Bluff in Session 7}
\label{app:E2}

In session 7, hand 16, Grok held A$\heartsuit$8$\clubsuit$ (ace-high with no pair and no draw) on a board that ran 2$\diamondsuit$ T$\diamondsuit$ 3$\spadesuit$ on the flop and 2$\clubsuit$ on the turn. DeepSeek held K$\clubsuit$J$\clubsuit$ and called the flop bet before folding to the turn barrel, leaving Grok to win a pot of \$260 as the last player standing.

\begin{table}[h]
\centering
\small
\caption{Hand trace: Session 7, Hand 16. Blinds \$5/\$10.}
\begin{tabular}{lllrl}
\toprule
Street & Player & Action & Amount & Key phrase \\
\midrule
Preflop & Grok     & raise   & 30  & MP open, A8o playable \\
Preflop & DeepSeek & call    & 30  & KJs, wide call \\
Preflop & (others) & fold    & 0   & (not shown) \\
Flop [2d Td 3s] & Grok     & bet     & 40  & ``air, fold equity'' \\
Flop   & DeepSeek & call    & 40  & backdoor draws, overcard \\
Turn [2c] & Grok     & bet     & 105 & ``barrel nits on paired board'' \\
Turn   & DeepSeek & fold    & 0   & ``large bet = strength'' \\
\bottomrule
\end{tabular}
\end{table}

Grok's flop reasoning explicitly acknowledged hand weakness (``Flop completely misses my Ah8c, ace-high no pair/draw'') and identified a favourable bluffing spot, namely a low dry board where the preflop raiser's range contains many strong hands and the c-bet is therefore credible. The turn barrel on the pairing deuce was justified by referencing DeepSeek's profile (``nits fold to repeated aggression''), which is a score-2 opponent-model demonstration under the M3 rubric. The hand contributes to Grok's M2 score of 0.83 and M3 of 0.46.

\subsection{Kimi's Hand-Strength Hallucination in Session 5}
\label{app:E3}

In session 5, hand 15, Kimi held J$\heartsuit$Q$\heartsuit$ and faced a board of 4$\heartsuit$ 2$\heartsuit$ 9$\diamondsuit$ 6$\heartsuit$ on the turn. With four hearts on the board and Kimi holding two hearts, the hand is a queen-high flush \textit{draw} (needing a fifth heart), not a made flush. Kimi's reasoning nevertheless stated that it held a ``queen-high flush'' and bet \$180 for value.

\begin{table}[h]
\centering
\small
\caption{Hand trace: Session 5, Hand 15, turn action. Kimi's stated hand
vs. actual hand.}
\begin{tabular}{lll}
\toprule
Field & Stated (in reasoning) & Actual \\
\midrule
Hand type & Queen-high flush & Flush draw (4/5 hearts) \\
Cards mentioned & A$\heartsuit$, K$\heartsuit$ & Not on board or in hole \\
Action & Bet \$180 (value) & Bet \$180 (as draw, not value) \\
Equity & ``ahead of most'' & $\approx 35\%$ vs. made pairs \\
\bottomrule
\end{tabular}
\end{table}

Kimi's full reasoning was: \textit{``With QhJh giving us a queen-high flush on the turn, we are ahead of most of the opponents' ranges that called the flop (lower flushes, sets, two pairs). A half-pot bet extracts value from those worse hands while limiting our loss if we face a raise from the rare higher flush (Ah or Kh). Checking would give free cards to full house draws and miss value.''}

This reasoning hallucinated Ah and Kh as ``higher flushes'' to fear even though neither card was on the board or in Kimi's hand, and it misidentified the draw as a made hand. The hand exemplifies the M8 hallucination component, since the card mentions (``Ah or Kh'') that are absent from both hole cards and board are flagged as hallucinated references. Kimi's M8 score of 0.5955 and M1 bet-sizing score of 0.3614 are both the lowest in the cohort, consistent with systematic mechanical errors.

\section{Implementation Notes}
\label{app:impl}

\subsection{Hand Evaluation}

Hand evaluation uses a five-of-seven hand evaluator that returns the canonical hand description; ties are resolved by the standard kicker comparator. The evaluator is widely used in JavaScript poker engines and exposes both a single-hand solver and a multi-hand winner selector that returns more than one hand when the strongest hands tie. Delegating to a vetted evaluator avoids hand-rolling a rank evaluator and guarantees correct split-pot accounting at showdown.

\subsection{Equity Calculation}

Run-out equity is estimated via a Monte-Carlo simulator with 1{,}000 random run-outs per decision, and the win fraction is recorded as the equity attribute attached to each action record. Monte Carlo sampling was chosen over an exhaustive enumerator because 1{,}000 samples are sufficient for the granularity that the metrics consume and remain fast enough to run in-line during the game. The equity field is absent from early sessions recorded before the equity-tracking pipeline was added and defaults to \texttt{null} in those cases.

\subsection{Side-Pot Accounting}

When players are all-in for different amounts, we compute a single main pot plus per-player contribution tracking rather than materialising a separate pot object for each layer. This keeps the data model simple: at showdown, winners are determined layer by layer, with the player eligible for each layer winning it if they hold the best hand among eligible players. The pot-reconstruction logic lives in the game engine. Per-player contribution tracking was preferred because it avoids the double-counting bugs that afflict naive layered-pot implementations when the same player goes all-in twice across rebuys, even though rebuys are not enabled in the arena configuration.

\subsection{Parse Fallback Ladder}

Model responses are processed by a structured-output parser; the JSON tool-call is parsed first, falling back to regex extraction of \texttt{ACTION} and \texttt{REASONING} fields, and finally to a default check/fold action when both fail. The first stage attempts a tool-call primary: the model is prompted with a poker-action tool schema, and a valid tool-call JSON is accepted directly. The second stage applies regex extraction over free-form text, catching responses where the model ignores the tool interface but still produces structured prose. The third stage substitutes a default action with a generic reasoning string. The ladder is structured this way because each stage is strictly more permissive than the previous one, so a response that passes an earlier stage is trusted more than a response that only passes a later one. All parse failures are logged and counted, and the fallback rate remains below 2\% across all models.

\subsection{Prompt Caching}

Provider-specific caching is used wherever available to reduce per-action cost and latency. Anthropic's ephemeral cache headers provide a 90\% discount on cached input tokens, Google's implicit server-side caching is transparent to the caller, xAI and OpenAI cache automatically at approximately a 50\% discount, Moonshot (Kimi) uses automatic caching, and Qwen does not support prompt caching. The system prompt and the memory blob are the primary cache candidates because they are static within a session, so caching them shifts most of the per-action cost onto the relatively small action-prompt tail.

\subsection{Agent Anonymisation}

Each session maintains a stable alias map built at session start. All opponent names in prompts are replaced by ``Player 1'', ``Player 2'', and so on up to ``Player $N$'', while the focal agent's own name is preserved so that it can recognise itself. The same alias is used for the same opponent across all sessions of a run so that cross-session memory notes remain consistent, and model identifiers such as \texttt{anthropic/claude-opus-4.6} are never exposed to any agent. Anonymisation is important because it prevents the agent from conditioning on brand priors rather than on observed play.

\subsection{Ablation Execution}

The ablation runner executes the six (model, condition) cells under matched random seeds and writes per-session chip deltas to a dedicated SQLite table (separate from the main tournament data) for paired analysis. Condition A uses pre-recorded game seeds from session 50 of the main dataset, ensuring that both conditions face identical board run-outs, and Condition B is identical to A except that the full persistent-memory mechanism is enabled. Sharing seeds across conditions is what allows the paired statistical tests in Appendix~\ref{app:D3} to be meaningful.

\end{document}